\documentclass[10pt,a4paper,twocolumn]{article}

\usepackage[T1]{fontenc}
\usepackage[utf8]{inputenc}
\usepackage[sc]{mathpazo}                 
\usepackage[scaled=0.90]{helvet}          
\usepackage[a4paper,top=2.1cm,bottom=2.2cm,left=1.75cm,right=1.75cm,columnsep=0.65cm,headsep=0.55cm]{geometry}
\usepackage{microtype}

\usepackage[table]{xcolor}
\definecolor{accent}{RGB}{0,92,118}       
\definecolor{accentlight}{RGB}{235,244,247}
\definecolor{inkgrey}{RGB}{85,95,105}

\usepackage{amsmath,amssymb,amsthm}
\usepackage{graphicx}
\usepackage{placeins}
\usepackage{booktabs,multirow,rotating}
\usepackage[super,sort&compress,comma]{natbib}
\usepackage{xurl}
\usepackage{enumitem}
\setlist[itemize]{leftmargin=1.1em,itemsep=2pt,topsep=4pt,label={\color{accent}\small$\blacktriangleright$}}
\usepackage{flafter}

\usepackage{titlesec}
\titleformat{\section}{\sffamily\bfseries\Large\color{accent}}{\thesection}{0.6em}{}[{\color{accent}\titlerule[0.8pt]}]
\titlespacing*{\section}{0pt}{16pt plus 3pt}{8pt}
\titleformat{\subsection}{\sffamily\bfseries\large\color{accent}}{\thesubsection}{0.5em}{}
\titlespacing*{\subsection}{0pt}{12pt plus 3pt minus 1pt}{4pt}
\titleformat{\paragraph}[runin]{\sffamily\bfseries\small\color{accent!80!black}}{}{0pt}{}[.]
\titlespacing*{\paragraph}{0pt}{6pt plus 2pt}{0.6em}

\usepackage{caption}
\DeclareCaptionLabelSeparator{bar}{\enspace\textcolor{accent}{|}\enspace}
\DeclareCaptionFont{accent}{\color{accent}}
\renewcommand{\figurename}{Fig.}
\usepackage{fancyhdr}
\fancypagestyle{first}{\fancyhf{}\fancyfoot[C]{\sffamily\footnotesize\color{inkgrey}\thepage}}

\usepackage[colorlinks=true,linkcolor=accent,citecolor=accent,urlcolor=accent]{hyperref}

\let\OrigFloatBarrier\FloatBarrier\renewcommand{\FloatBarrier}{}             
\newcommand{\model}{Foehn}

\newcommand{\pmtf}{PM$_{2.5}$}

\newcommand{\edfig}{Extended Data Fig.}

\DeclareRobustCommand{\rev}[1]{#1}
\DeclareRobustCommand{\nrev}[1]{#1}

\hypersetup{pdftitle={Global probabilistic forecasting of air pollution at the kilometre scale},
  pdfauthor={Ammar Kheder, Michael Boy, Zhi-Song Liu}}

\begin{document}

\twocolumn[{%
\thispagestyle{first}
\begin{flushleft}
{\sffamily\footnotesize\bfseries\color{accent}\MakeUppercase{Preprint}\quad\textcolor{inkgrey}{\mdseries Atmospheric science \textperiodcentered{} Machine learning}}\par
\vspace{8pt}
{\color{accent}\rule{\textwidth}{1.4pt}}\par
\vspace{12pt}
{\sffamily\bfseries\fontsize{23}{28}\selectfont Global probabilistic forecasting\\[2pt] of air pollution at the kilometre scale\par}
\vspace{14pt}
{\sffamily\large Ammar Kheder$^{1,2,3,*}$\enspace\textperiodcentered\enspace Michael Boy$^{1,2,3}$\enspace\textperiodcentered\enspace Zhi-Song Liu$^{1,2,3}$\par}
\vspace{8pt}
{\sffamily\footnotesize\color{inkgrey}
$^1$Department of Computational Engineering, LUT University, Lahti, Finland\\
$^2$Institute for Atmospheric and Earth System Research, University of Helsinki, Helsinki, Finland\\
$^3$Atmospheric Modelling Centre Lahti (AMC-Lahti), Lahti, Finland\\
$^*$Corresponding author: \href{mailto:ammar.kheder@lut.fi}{ammar.kheder@lut.fi}\par}
\vspace{14pt}
\colorbox{accentlight}{\parbox{\dimexpr\textwidth-2\fboxsep\relax}{%
\vspace{9pt}%
\hspace*{8pt}\parbox{\dimexpr\textwidth-2\fboxsep-16pt\relax}{%
{\sffamily\bfseries\small\color{accent}Abstract}\par\vspace{4pt}
\small\linespread{1.08}\selectfont Fine particulate matter (\pmtf{}) refers to particulate matter with an aerodynamic diameter of 2.5\,\textmu m or less.
\nrev{It comes from human activities, such as traffic and industry, and from wildfires, desert dust and gases emitted by vegetation\cite{mcduffie2021source}.}
\nrev{Short-term exposure to it accounts for about a million deaths a year\cite{yu2024short}.}
\nrev{Operational forecasts run on grids tens of kilometres wide\cite{peuch2022cams}, which hide the pollution breathed by nearly two hundred million people a day.}
Machine learning has transformed weather forecasting\cite{lam2023graphcast,price2025gencast}\nrev{, but data-driven forecasts of air pollution still blur extremes and lose skill as the air travels.}
Here we introduce Foehn, a generative model that learns from \nrev{global kilometre-scale estimates} of \pmtf{}\cite{wei2023global} and computes where the air comes from.
\nrev{Over a year of retrospective forecasts, it outperforms the operational forecast at three stations in four and \nrev{forecasts at the scale of this hidden exposure}.}
\nrev{Its ensembles are calibrated, and their mean detects local pollution extremes four days ahead as well as the operational forecast does one day ahead.}
\nrev{Such forecasts could warn the millions whom coarse maps miss, once kilometre-scale data are available in real time.}
\par}%
\vspace{9pt}}}\par
\vspace{18pt}
\end{flushleft}
}]



In January 2022, haze from human activities covered the North China Plain for a week, and in July 2022 smoke from forest fires in south-western France drifted over a bird reserve.
Such episodes matter because fine particulate matter (\pmtf{}), particles with an aerodynamic diameter of 2.5\,\textmu m or less, is one of the leading risk factors for premature death worldwide\cite{gbd2020risk}.
It is emitted directly or formed in the air from gases released by human activities, such as traffic, industry, power generation, domestic heating and agriculture, and by natural sources, such as wildfires, desert dust and vegetation, whose volatile organic compounds form secondary organic aerosol\cite{mcduffie2021source,hallquist2009soa}.
Short-term exposure, over hours to days, accounts for about a million deaths a year\cite{yu2024short}.
\nrev{On this timescale, forecasts and warnings allow people to protect themselves, for example by limiting time outdoors or avoiding polluted neighbourhoods.}
The harm also extends beyond people: wildfire smoke causes respiratory distress and changes in behaviour in wild animals, and birds repeatedly exposed to smoke lose weight\cite{sanderfoot2021review,nihei2024wildfire}.
Global operational forecasts, such as those of the Copernicus Atmosphere Monitoring Service (CAMS)\cite{peuch2022cams}, come from chemistry-transport models run on grids approximately 40\,km wide.

These forecasts have four limitations.
First, a grid this coarse hides exposure that varies greatly over short distances.
Combustion pollutants vary up to eightfold within a single city block\cite{apte2017high}.
Second, such a grid also hides inequality: in many cities, poorer and minority neighbourhoods breathe more polluted air than their neighbours\cite{hajat2015socioeconomic,tessum2021pm25}.
Third, measurements cannot fill these gaps: in 2019, 141 of 234 countries had no regular \pmtf{} monitoring, and the average person lived 220\,km from the nearest monitor\cite{martin2019no}.
Finally, each forecast is a single deterministic field, with no probability on which a warning can act.
Reaching people at the scale of a neighbourhood, with reliable probabilities, calls for another approach.

Since 2023, \nrev{neural networks} have matched or surpassed physics-based weather forecasting\cite{bi2023pangu,lam2023graphcast,price2025gencast}.
\nrev{Deep learning has also been applied to air pollution forecasting}, from national networks of stations\cite{wang2020pm25gnn,liang2023airformer} to global forecasts\cite{bodnar2025aurora,gui2026aigamfs,harder2026aifscompo}.
\nrev{However, these models are trained to minimize a pointwise error, such as the mean squared error, which smooths the forecasts in space and time and blurs the extremes that matter for health.}
We refer to this problem as oversmoothing.
Furthermore, most \nrev{neural network} models step forward through their own output, a few hours to a day at a time (a rollout), so that errors accumulate \nrev{over long lead times}\cite{bi2023pangu,lam2023graphcast,bodnar2025aurora}.
Pollution also travels with the air, so that after several days the concentration at a place depends less on its initial state than on where its air came from and which sources it passed over. We refer to this second problem as forecast drift.

In this paper, we present Foehn, a generative neural network that forecasts daily \pmtf{} at $1\,\mathrm{km}\times1\,\mathrm{km}$ resolution over all inhabited land up to 7 days ahead.
Foehn first computes where the air arriving at each place comes from via a particle model driven by the forecast wind.
It then passes the learned tracers to the flow matching, a generative method that produces plausible future fields via stochastic sampling.
Over the test year 2022, its error 1 day ahead at 6,324 monitoring stations, which is 17\% lower than that of the operational CAMS forecast.
Specifically, \nrev{this study} provides:
\begin{itemize}\setlength{\itemsep}{1pt}
\item \nrev{kilometre-scale map of exposure, revealing the 186 million people a day above a World Health Organization interim target whom 40\,km grids hide;}
\item kilometre-scale forecasts of daily \pmtf{} over all inhabited land, outperforming the CAMS forecast at 75\% of 5,954 stations;
\item \nrev{calibrated ensembles whose mean detects local pollution extremes 4 days ahead as well as the CAMS forecast does 1 day ahead;}
\item a computed transport that adds skill in the first 3 days, and emission inputs that lower the probabilistic error by 6--8\% from 2 to 7 days ahead;
\item \nrev{kilometre-scale ensembles of wildfire smoke, placing it over a nature reserve 5 days ahead.}
\end{itemize}

\subsection*{The Foehn forecasting system}

\begin{figure*}[!tp]
\centering\includegraphics[width=\textwidth]{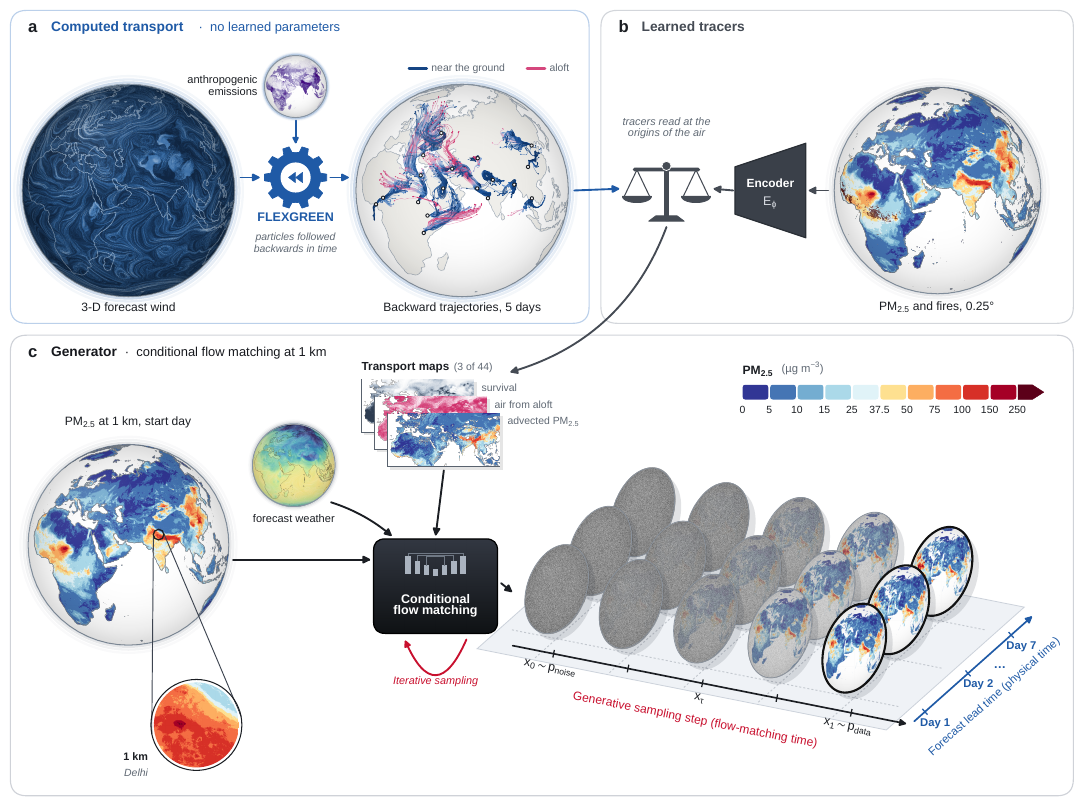}
\caption{\textbf{Foehn computes where the air comes from and learns what it carries.}
\textbf{a}, Computed transport, without learned parameters. The three-dimensional forecast wind (shown at 850\,hPa, 00\,UTC on 20 January 2022, the start day) drives FLEXGREEN. FLEXGREEN releases particles in every 0.5$^\circ$ land cell over the target day and follows them back to the start day. \nrev{On the way, it sums the emissions that the air takes up: the anthropogenic emissions (purple; black carbon shown) and those of the fires of the start day.} Lines, 5-day back-trajectories of the air reaching 18 cities on 25 January 2022, blue near the ground and pink aloft.
\textbf{b}, Learned tracers. An encoder turns the start day (\pmtf{} at 0.25$^\circ$ and fires) into 32 tracers. Each place reads them where its air was on the start day, in the mixed layer.
\textbf{c}, Generator. A conditional flow-matching network turns a noise field into a 1\,km forecast in four sampling steps (red axis), for each lead time (blue axis). It is conditioned on the start day at 1\,km (lens: Delhi), the forecast weather (2-m temperature and wind shown) and 44 transport maps: the tracers of \textbf{b}, the emission terms of \textbf{a}, the share of air from aloft, its survival and the advected \pmtf{} (three shown, 5 days ahead). Discs, one member of the forecast issued on 20 January 2022 for 21, 22 and 27 January, from noise ($x_0$) to forecast ($x_1$). The three intermediate discs blend the two at the flow times reached after the first three steps (0.03, 0.17 and 0.53). Each noise field gives one ensemble member. Colour scale, daily \pmtf{} (\textmu g\,m$^{-3}$).}
\label{fig:schematic}
\end{figure*}

\nrev{From the state of the air on a start day and a weather forecast, Foehn produces an ensemble of plausible 1\,km maps for each of the next 7 days.}
It learns from the Global High Air Pollutants dataset (GHAP)\cite{wei2023global}, which gives daily \pmtf{} at 1\,km over land from 2017 to 2022.
These data are derived from satellite observations and meteorological reanalyses and fitted to station measurements, and we call them the 1\,km estimates.
Foehn has three parts (Fig.~\ref{fig:schematic}): (1) FLEXGREEN, which computes where the air arriving at each place was on the start day and the emissions it takes up on its way; (2) learned tracers, which this air carries from its origins; and (3) a conditional generative flow matching, which produces the 1\,km$\times$1\,km forecasts.

The first two parts describe how pollution moves.
FLEXGREEN re-implements on graphics processors the physics of the backward mode of the FLEXPART particle dispersion model\cite{stohl2005flexpart,pisso2019flexpart}.
For every 0.5$^\circ$ land cell, it releases particles of air on the target day and follows them backwards through the three-dimensional forecast wind, with turbulence, deposition and removal by rain, to where they were on the start day (Methods).
The result, a map of where the air of each place came from, is called the Green's function of the transport.
On the way, FLEXGREEN also sums the fire emissions of the start day\cite{kaiser2012gfas} and the emissions from human activities\cite{soulie2024cams} that the air takes up near the ground.

Because transport is linear, these paths can carry any quantity.
A small neural network therefore turns the start-day maps of \pmtf{} and fires into 32 quantities, which each place reads where its air was on the start day, keeping only the air that was in the mixed layer, the lowest layer of the atmosphere, in contact with the surface.
These learned quantities are called latent tracers.
Whereas FLEXGREEN has no learned parameters and is computed once, the tracers are trained together with the generator: the network chooses what to transport, and the physics decides where it comes from.

The third part generates the forecast.
A neural network learns to turn random noise into a plausible map of daily \pmtf{} at 1\,km in a few steps, a method called conditional flow matching\cite{lipman2023flow,liu2023flow}.
\nrev{It is conditioned on the start day, its fires and its weather, on the weather forecast for the target day and on the transport maps, which carry the latent tracers and the emissions (Fig.~\ref{fig:schematic}c).}
Running it from different noise fields gives an ensemble whose spread measures the uncertainty of the forecast.
Flow matching alone gives members that spread about half as much as their error.
A short fine-tuning on the continuous ranked probability score (CRPS), which rewards both accuracy and calibration, brings the spread up to the error (Methods).
\nrev{Concretely, Foehn has 14 million learned parameters and is trained on 2017--2020, validated on 2021 and tested on 2022. In total, training the final model took 25.8\,h on 32 graphics processing units.}

This design addresses both problems.
Against oversmoothing, the network learns the distribution of \pmtf{} rather than its average, so that each member keeps fine detail and high values.
In a controlled benchmark over Europe, its forecasts are closer to the measurements and sharper on the peaks than those of five networks trained by squared-error regression on identical inputs.
They are also more accurate than the CAMS forecast and Aurora\cite{bodnar2025aurora}, a foundation model of the Earth system (Extended Data Figs.~\ref{fig:speccost}--\ref{fig:eu_extremes}).
Against forecast drift, the transport is computed rather than learned, and each lead time is forecast directly from the start day, so that errors cannot build up by feeding forecasts back as inputs (Fig.~\ref{fig:schematic}c).


\FloatBarrier
\subsection*{\nrev{Hidden exposure and public health}}

\nrev{Air-quality warnings protect public health only if they reach the people who breathe polluted air.}
A 40\,km grid averages a city with its surroundings\nrev{, whereas exposure differs between neighbourhoods and between social groups\cite{hajat2015socioeconomic,tessum2021pm25}.}
\nrev{Measuring the exposure that such a grid hides is difficult for two reasons.}
First, exposure depends on where people live within a 40\,km cell.
\nrev{Second, the \nrev{estimates} that map it at 1\,km have errors and regional biases (Methods).}

\begin{figure*}[!tp]
\centering\includegraphics[width=0.82\textwidth]{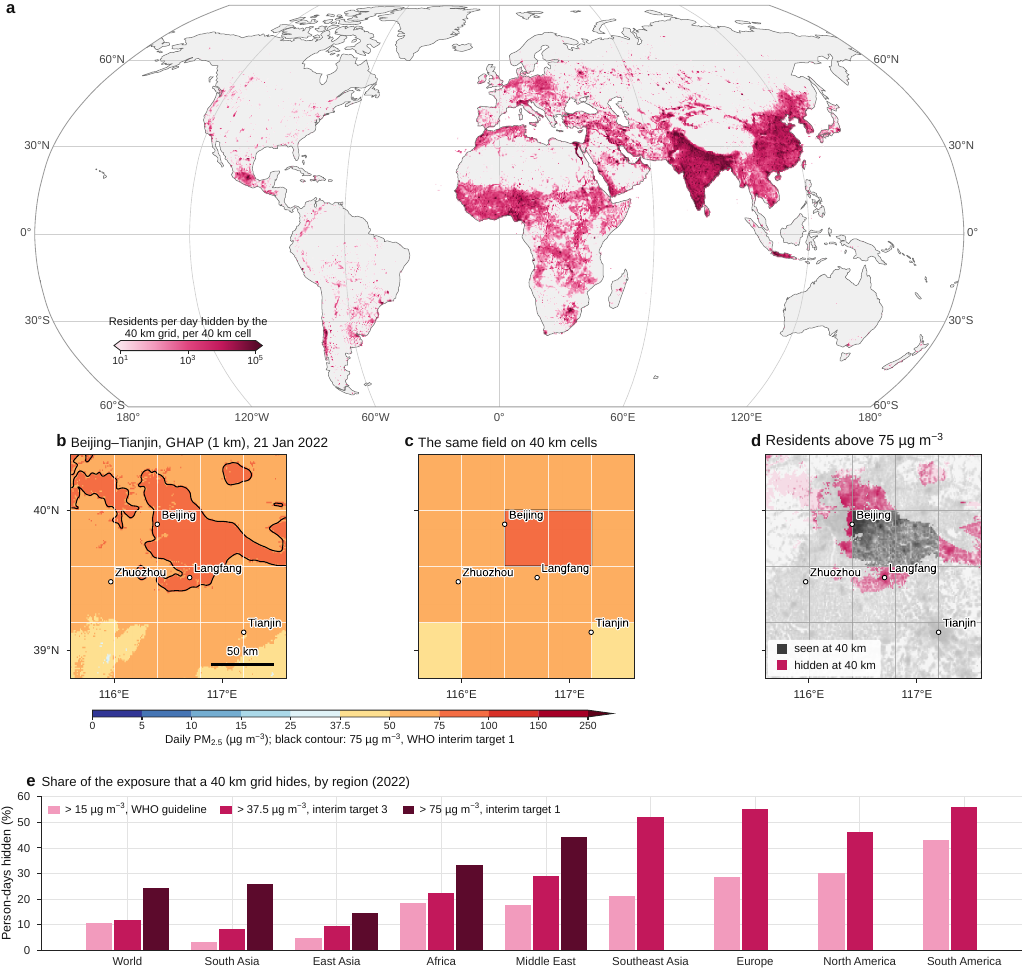}
\caption{\textbf{\nrev{Even a perfect 40\,km forecast cannot warn millions of people a day who breathe polluted air.}} \textbf{a}, Number of residents per day in 2022 whose 1\,km grid cell was above the World Health Organization (WHO) 24-h interim target 3 (37.5\,\textmu g\,m$^{-3}$) while the mean of the surrounding 40\,km cell was below it, mapped per 40\,km cell: people whom even a perfect 40\,km forecast could not warn (\nrev{1\,km estimates} of the Global High Air Pollutants dataset, GHAP, and population, without any forecast). \textbf{b}--\textbf{d}, Beijing--Tianjin on 21 January 2022, the most polluted day of the year for the world's population: daily \pmtf{} in GHAP with the contour of 75\,\textmu g\,m$^{-3}$, WHO interim target 1 (\textbf{b}); the same field averaged over 40\,km cells, the size of the cells of the CAMS forecast, which is what a perfect 40\,km forecast gives (\textbf{c}); and the residents above 75\,\textmu g\,m$^{-3}$ seen (grey) or hidden (pink) by the 40\,km cells (\textbf{d}). \nrev{\textbf{e}, Of the person-days of 2022 above each threshold in the residents' 1\,km grid cell, the share in 40\,km cells whose mean is below it, by region: the WHO 24-h guideline level (15\,\textmu g\,m$^{-3}$; light) and interim targets 3 (37.5\,\textmu g\,m$^{-3}$) and 1 (75\,\textmu g\,m$^{-3}$; dark); same data as \textbf{a}. A bar is shown when the region has at least one billion person-days above its threshold.}}
\label{fig:hidden}
\end{figure*}

On 21 January 2022, nearly half of the people of the Beijing--Tianjin area who breathed air above 75\,\textmu g\,m$^{-3}$ lived in 40\,km cells below that level (Fig.~\ref{fig:hidden}b--d).
\nrev{On the 40\,km cells of the operational grid, which are what a perfect 40\,km forecast would give, their exposure was averaged with that of cleaner surroundings.}
\nrev{This case shows that a coarse grid can hide nearly half of the people exposed to a severe episode, even around a capital city.}

\FloatBarrier
\subsection*{Kilometre-scale forecasts}

\nrev{Because \pmtf{} varies within a city, warnings help only if they reach the neighbourhoods in which the air is polluted.}
\nrev{Two reference systems, the CAMS forecast\cite{peuch2022cams} and Aurora\cite{bodnar2025aurora}, forecast \pmtf{} on grids of about 40\,km.}

\begin{figure*}[!tp]
\centering\includegraphics[width=\textwidth]{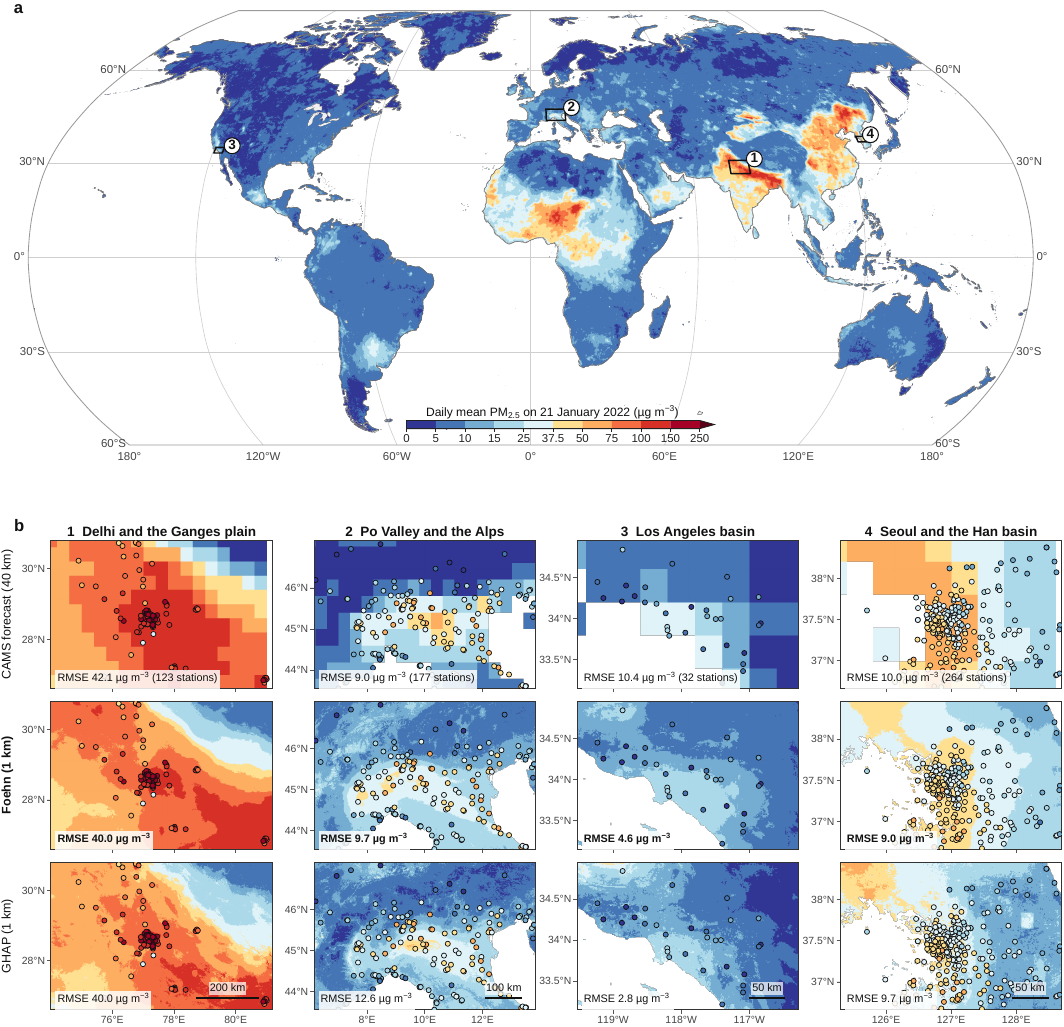}
\caption{\textbf{Foehn forecasts all inhabited land at 1\,km and resolves basins that the operational grid blurs.}
\textbf{a}, Foehn's forecast of daily \pmtf{} (ensemble mean of four members) issued at the end of 20 January 2022 for 21 January,
the day of 2022 with the highest population-weighted \pmtf{}. \textbf{b}, The forecast of \textbf{a} in four polluted basins, each
monitored by a different official network (1, Delhi, Central Pollution Control Board of India; 2, Po Valley, European Environment
Agency; 3, Los Angeles, US Environmental Protection Agency; 4, Seoul, AirKorea). Shown are the CAMS forecast issued at the same
time (00\,UTC run of 21 January, 0.4$^\circ$ grid), Foehn at 1\,km and the \nrev{1\,km estimates} (GHAP, made after the fact, not
a forecast), with the measurements of that day at the stations in the same colours. Numbers, root-mean-square error (RMSE) over
2022 at the stations of each basin, 1 day ahead, on identical station-days. Aurora and the error against the \nrev{estimates} in
four regions are shown in \edfig~\ref{fig:regions}.}
\label{fig:overview}
\end{figure*}

We combine the daily \nrev{1\,km estimates} of 2022 with the 1\,km population map of 2020, independently of any forecast.
A resident is then hidden when the \pmtf{} of their 1\,km grid cell exceeds a threshold and the mean of their 40\,km cell does not (Methods).
\nrev{The thresholds are the 24-h guideline level and interim targets of the World Health Organization (WHO).}

On an average day of 2022, 186 million people breathed air above WHO interim target 3 (37.5\,\textmu g\,m$^{-3}$) while the mean of their 40\,km cell was below it, especially in Africa and the Middle East (Fig.~\ref{fig:hidden}a,e).
Where terrain confines a city, the 40\,km mean can fall below a threshold that most of the city exceeds, as in Santiago de Chile and in the Nile valley around Cairo (\edfig~\ref{fig:exposure}a--f).
\nrev{Foehn forecasts at this scale, and its maps can support more equitable exposure assessments and public-health decisions, in particular in regions with few data.}
\nrev{These numbers rest on the \nrev{estimates} and inherit their errors.}

\FloatBarrier
\subsection*{Extreme pollution episodes}

Early warnings matter the most during pollution extremes, and a warning must give the probability of such an episode.
\nrev{The CAMS forecast and Aurora give one deterministic field and no calibrated probability of a local extreme.}
We identify three challenges in extreme forecasting.
First, an extreme is local: a day that is ordinary in northern China could be severe in northern Europe.
Second, training on a pointwise error averages away the peaks\cite{li2026aurora}.
Finally, a probability can be trusted only if the spread of the ensemble matches its error, so that an event forecast with a probability of 30\% occurs about three times in ten.

\begin{figure*}[!tp]
\centering\includegraphics[width=\textwidth]{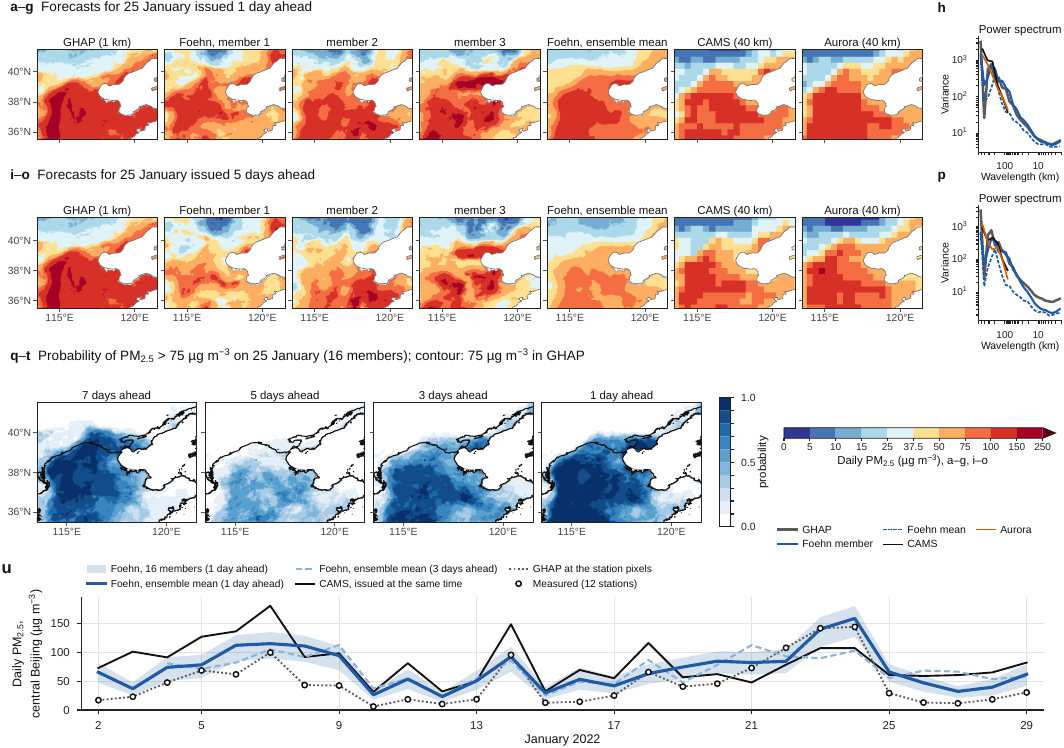}
\caption{\textbf{Each member of Foehn's ensemble keeps the sharp structure of a haze episode that the operational forecasts smooth
away.} Daily \pmtf{} on 25 January 2022, the peak of the haze over the North China Plain. \textbf{a}--\textbf{g}, \nrev{The 1\,km estimates (\textbf{a}) and the forecasts issued 1 day ahead:} three of Foehn's 16 members (\textbf{b}--\textbf{d}), Foehn's ensemble mean
(\textbf{e}) and the CAMS (\textbf{f}) and Aurora (\textbf{g}) forecasts issued at the same time, on their 0.4$^\circ$ grid.
\textbf{h}, Variance per decade of wavelength of each field over the panel area (radially averaged power spectrum).
\textbf{i}--\textbf{o}, \textbf{p}, The same, issued 5 days ahead. \textbf{q}--\textbf{t}, Probability of exceeding
75\,\textmu g\,m$^{-3}$ (WHO 24-h interim target 1) from Foehn's 16 members, issued 7, 5, 3 and 1 days ahead; black contour,
75\,\textmu g\,m$^{-3}$ in the \nrev{estimates}. \textbf{u}, Daily mean (UTC day) of the measurements at the 12 monitors of the China
National Environmental Monitoring Centre (CNEMC) in central Beijing in January 2022 (circles), GHAP at their grid cells (grey
dotted) and the forecasts issued 1 day earlier: Foehn's 16 members (band, minimum to maximum; line, ensemble mean) and the CAMS
forecast issued at the same time (black); light blue dashed, Foehn's ensemble mean issued 3 days earlier.}
\label{fig:realism}
\end{figure*}

\nrev{We evaluate the forecasts of 2022 against the 1\,km estimates on identical land cells and days and, because these estimates have biases of their own, against measurements from 6,324 stations of six official networks that were never used for training} (Methods). 
\nrev{Foehn's ensembles have four members, and \edfig~\ref{fig:regions} compares the three systems over four regions on one day.}

Over all land, Foehn follows the \nrev{1\,km estimates} closely: their correlation is 0.89 at a lead time of 1 day and 0.80 at 4 days.
For the CAMS forecast and Aurora, it is approximately 0.6, and both are biased low relative to these data.
One day ahead, Foehn's error is a third below that of each place's climatology (the mean of the \nrev{estimates} of 2017--2020 within a week of the same calendar day).
After a week, its error is only 3\% below that of the climatology (9.9 against 10.2\,\textmu g\,m$^{-3}$; correlation 0.76 against 0.75), as the predictability of the weather fades.

\nrev{At the measurement stations,} Foehn's root-mean-square error (RMSE) 1 day ahead is 12.7\,\textmu g\,m$^{-3}$, against 15.2 for the CAMS forecast and 15.7 for Aurora (\edfig~\ref{fig:fair}).
Four days ahead, the gains are 12\% and 17\%.
Foehn is more accurate than the CAMS forecast at 75\% of the 5,954 stations with at least 30 common forecast days.
The gain is as large at the 4,295 stations of the five networks outside Europe.
\nrev{From ensemble forecasting,} its CRPS 1 day ahead, 4.3\,\textmu g\,m$^{-3}$, is about half that of the CAMS forecast (7.9) and Aurora (8.3).
Seven days ahead, at 5.3. It remains well below that of the climatology at the measurement stations (9.7).

This comparison has a limit.
The stations are not independent of the gridded global \nrev{estimates}, which were fitted to measurements from about 9,500 stations, including the European, Chinese and US networks\cite{wei2023global}.
The accuracy of these data also varies widely between networks.
\nrev{They match the Chinese network almost exactly (RMSE 2.4\,\textmu g\,m$^{-3}$) but are no better than their own climatology at the European network (11.2 against 11.0) and 5.9\,\textmu g\,m$^{-3}$ too low at the Indian network (Methods).}
Foehn's gain is largest in China, where both reference models overestimate the haze events.
In India, Foehn is only as accurate as the CAMS forecast and Aurora, and it \nrev{amplifies the low bias of the estimates ($-$10.9 against $-$5.9\,\textmu g\,m$^{-3}$)}.

At the European and Japanese networks, its forecast 1 day ahead is closer to the measurements than the \nrev{estimates} of the same day (RMSE 9.8 against 11.2, and 5.9 against 6.6; \edfig~\ref{fig:fair}a--f).
\nrev{We hypothesize that averaging the members acts as a filter.
The mean keeps the part of the \nrev{estimates} that the start day and the weather can predict, and averages out the rest.
At these networks, this rest would be partly error in the \nrev{estimates}, which would explain why the mean is closer to the measurements than the \nrev{estimates} themselves.}

The gain is not a matter of resolution alone: averaged to 0.5$^\circ$ before being read at the stations, Foehn's error is still below that of both reference systems.
At 1\,km, it is 5\% below that of its own 0.25$^\circ$ average (\edfig~\ref{fig:fair}g).
\nrev{Foehn's design} also does not depend on the \nrev{1\,km estimates}.
Retrained in Aurora's own protocol, on the coarse CAMS analyses and without any 1\,km or station data, the same design outperforms Aurora by 9\% in RMSE 3 days ahead and 12\% 5 days ahead (Methods; Extended Data Figs.~\ref{fig:auroraset} and \ref{fig:aurorareg}).

The day of 2022 with the highest population-weighted \pmtf{} was 21 January.
The forecast issued the evening before placed the haze over the Indo-Gangetic Plain, the North China Plain and the Sahel (Fig.~\ref{fig:overview}a).
At 1\,km, \nrev{it resolved details that} the 40\,km grid of CAMS missed, such as the edge of the Indo-Gangetic Plain, the Po Valley along the Alps, the Los Angeles basin and the hills around Seoul.
\nrev{The station measurements, plotted in the same colours, showed the same patterns} (Fig.~\ref{fig:overview}b).
This case shows that the network resolves pollution at the scale of a basin over the whole globe.

\nrev{Foehn's ensemble mean} detects the days above each place's own 99th percentile better than the CAMS forecast and Aurora in every region (fewer false alarms).
Four days ahead, it detects them as well as the CAMS forecast does 1 day ahead (\edfig~\ref{fig:exposure}g,h).
At the measurement stations, this holds with the same thresholds, but not beyond 1 day with each station's own percentiles (Methods).
The proposed particle transport contributes: without it, the same network detects fewer extremes in every region 3 days ahead.
For a user whose protection costs a twentieth of the loss that it avoids, Foehn's warnings have more than twice the potential value of those of both reference systems, a value estimated retrospectively (\edfig~\ref{fig:exposure}i,j).
Foehn's ensemble mean, however, is smoother than its members, and the peak of an episode can be missed several days ahead.

On 25 January 2022, \nrev{a week-long haze episode} over the North China Plain reached its peak.
Each member of the ensemble issued 1 day ahead was a sharp field in which the haze filled the plain, stopped at the mountains and thinned towards the coast, as in the \nrev{estimates} (Fig.~\ref{fig:realism}a--g).
At the monitors of central Beijing, the ensemble captured the two days of haze (measured 142 and 144\,\textmu g\,m$^{-3}$; forecast 1 day earlier 140 and 159) and the clearing that followed, although it underestimated how far concentrations fell (Fig.~\ref{fig:realism}u).
Issued 1, 3 or 7 days ahead, the ensemble gave a high probability of exceeding 75\,\textmu g\,m$^{-3}$ (WHO interim target 1), where the \nrev{estimates} exceeded it, and a low one elsewhere.
However, from five days ahead, it under-forecast the peak (Fig.~\ref{fig:realism}q--t).
This case shows that calibrated ensembles turn sharp members into probabilities on which a warning can act.

\FloatBarrier
\subsection*{Transport and emissions}

\nrev{A} forecast several days ahead must know where the arriving air comes from and which emission sources it passes over.
Here we demonstrate that the computed transport and the emissions \nrev{taken up} along the way \nrev{add to} Foehn's skill from the first day to the seventh, \nrev{the first at short lead times and the second at longer ones}.
Learned forecasting models generally infer the transport from their inputs and accumulate errors through a rollout\cite{bi2023pangu,bodnar2025aurora}, whereas chemistry-transport models compute it on grids tens of kilometres wide\cite{peuch2022cams}.
\nrev{To separate the two contributions, we compare the base model, which has the transport inputs but not the emissions, with the same network trained without the transport inputs and with the same model fine-tuned with the emissions (Methods).}

\begin{figure*}[!tp]
\centering\includegraphics[width=\textwidth]{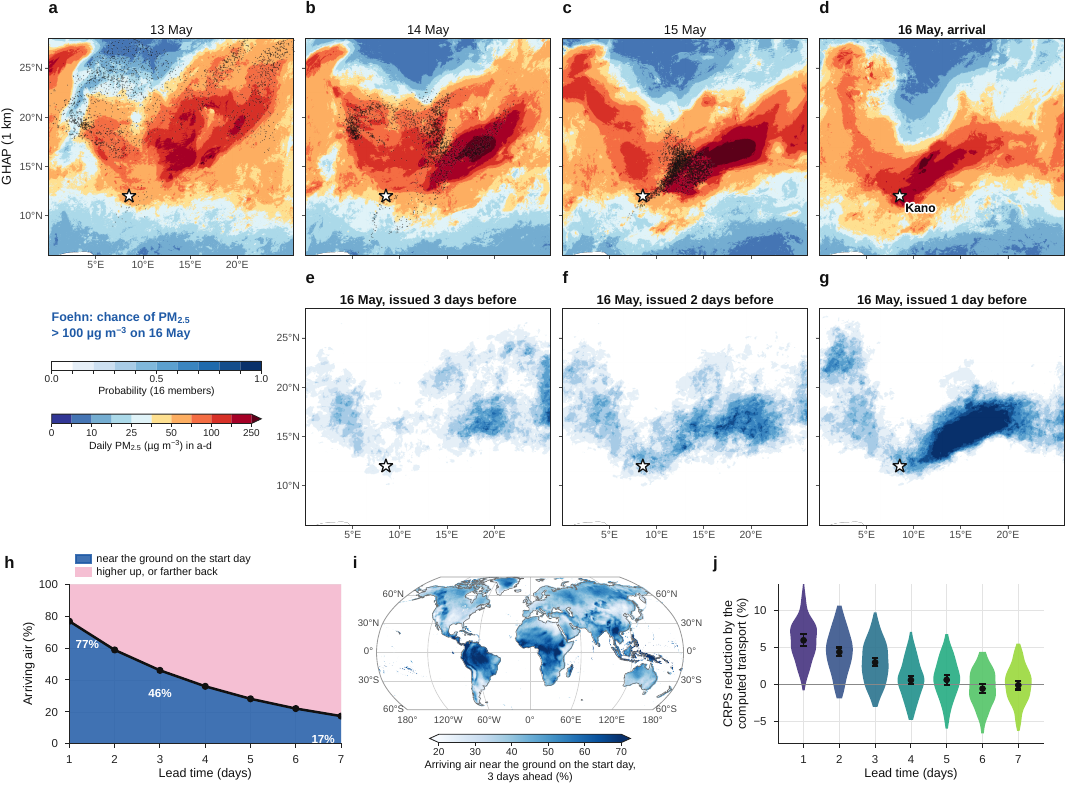}
\caption{\textbf{The computed transport traces a dust episode to its origin, and its gain fades as the start day loses influence.}
\textbf{a}--\textbf{d}, GHAP daily \pmtf{} at 1\,km over the central Sahel on 13--16 May 2022 and, on each of the first 3 days, where
the air reaching Kano (star) on 16 May was then (black dots), from the FLEXGREEN Green's functions of the forecasts started on 13
(\textbf{a}), 14 (\textbf{b}) and 15 May (\textbf{c}). \textbf{e}--\textbf{g}, Foehn's probability of \pmtf{} above
100\,\textmu g\,m$^{-3}$ on 16 May at 1\,km, from its 16 members, in the forecasts issued 3 (\textbf{e}), 2 (\textbf{f}) and 1 day
(\textbf{g}) before; compare with \textbf{d}. \textbf{h}--\textbf{j}, Why the gain of the computed transport fades with lead time
(all land, 2022). \textbf{h}, Where the air arriving on the target day was on the start day: in the mixed layer (blue) or higher up
or farther back (pink), by lead time. \textbf{i}, The same share 3 days ahead, place by place: where the start day still matters.
\textbf{j}, Reduction of the continuous ranked probability score (CRPS) against the \nrev{1\,km estimates} brought by the computed
transport: violins, for each start day; dots and bars, over all forecasts, with 95\% intervals.}
\label{fig:transport}
\end{figure*}

\nrev{The proposed particle transport shows that} the share of the arriving air that was \nrev{in the mixed layer} on the start day falls with lead time.
Over all land, it is approximately three-quarters for the next day, less than half after 3 days and a sixth after a week (Fig.~\ref{fig:transport}h,i).
The start day therefore carries progressively less information about the target day.
The gain from the computed transport declines accordingly: it lowers the CRPS by approximately 6\% 1 day ahead, by half as much 3 days ahead and not significantly beyond (Fig.~\ref{fig:transport}j).

\begin{figure*}[!tp]
\centering\includegraphics[width=\textwidth]{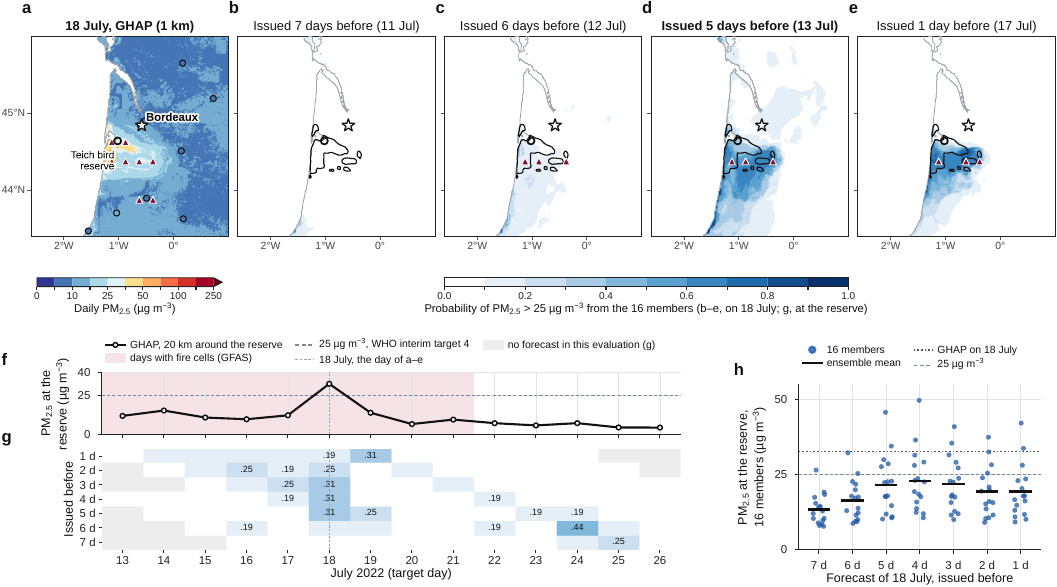}
\caption{\textbf{Five days ahead, Foehn places the smoke of the 2022 Gironde fires over a bird reserve.} The La Teste-de-Buch and
Landiras fires, which started on 12 July 2022, burnt about 18,000 hectares in the Gironde that month\cite{sanmiguel2023fires}; the La Teste fire burnt beside
the R\'eserve Ornithologique du Teich (open circle), a bird reserve on the Arcachon basin; star, Bordeaux. \textbf{a}, GHAP daily \pmtf{} at 1\,km on 18
July, the smokiest day at the reserve; triangles, the fire cells of the day in the Global Fire Assimilation System (GFAS); circles,
the European Environment Agency (EEA) stations, coloured by their measurement. \textbf{b}--\textbf{e}, Foehn's probability of
\pmtf{} above 25\,\textmu g\,m$^{-3}$ (WHO interim target 4) on 18 July, from its 16 members, issued 7 (\textbf{b}, 11 July, before
the fires ignited), 6 (\textbf{c}), 5 (\textbf{d}) and 1 day (\textbf{e}) before; triangles, the fire cells in the inputs of each
forecast; black contour, 25\,\textmu g\,m$^{-3}$ in GHAP on 18 July. \textbf{f}, GHAP daily \pmtf{} over the 20\,km surroundings of
the reserve, 13--26 July (shading, days with fire cells over the two fires; dashed, 25\,\textmu g\,m$^{-3}$). \textbf{g}, Foehn's
probability of exceeding 25\,\textmu g\,m$^{-3}$ over the same surroundings by target day (columns) and lead time (rows); numbers,
probabilities of 0.19 and above; grey, no forecast in this evaluation. \textbf{h}, The 16 members over the same surroundings for 18
July, by lead time (bars, ensemble mean; dotted, GHAP; dashed, 25\,\textmu g\,m$^{-3}$).}
\label{fig:smoke}
\end{figure*}

We therefore define extremes relative to each place's own climatology, as has been done for weather extremes\cite{liu2026tailweather}.
\nrev{A cell and day is labelled moderate, severe or extreme when it exceeds the local 90th, 95th or 99th percentile over the reference period 2017--2020 (Methods).}
We score their detection by the symmetric extremal dependence index (SEDI), a score that does not depend on how rare the events are\cite{ferro2011extremal}, and the warnings by their relative economic value (Methods).

\nrev{Each member of the ensemble is a possible scenario of a haze episode, with its own fine-scale structure.}
At scales finer than 20\,km, the members have as much spatial variability as the \nrev{1\,km estimates} 1 and 3 days ahead, and at least half of it after a week.
By contrast, the 40\,km forecasts of CAMS and Aurora have almost none below approximately 80\,km (Fig.~\ref{fig:realism}h,p).
\nrev{After the fine-tuning on the CRPS, the ensembles are calibrated against the estimates:} in 2022, their spread matches their error at every lead time (ratio 0.96--0.99; Methods), and on the validation year 2021 the ratio is 0.82--0.96 (\edfig~\ref{fig:fair}h).

\nrev{The emissions take over at longer lead times, provided that the inventory captures the dominant local sources} (\edfig~\ref{fig:emislead}).
Where they help, their gain does not fade.
In Europe, North America and South America, it rises over the first 2--4 days and then holds for the rest of the week, at approximately 7\%, 11\% and 4\% of the RMSE of the ensemble mean, respectively.
In East Asia, the two contributions cross at 4 days (\edfig~\ref{fig:emislead}a--d).
Over all regions together, the emissions lower the CRPS by 6--8\% from 2 to 7 days ahead.

The emissions do not help everywhere for two reasons (\edfig~\ref{fig:emislead}e,f; Methods).
First, the inventory misses the dominant sources of some regions.
Across most of Africa, desert dust and fires produce most of the \pmtf{}\cite{mcduffie2021source}, and global inventories misrepresent many local sources, such as fuelwood, charcoal and generators\cite{marais2016dice}.
There, the emissions degrade the ensemble mean by approximately 4\% at every lead time, and the CRPS improves at long lead times only because the ensemble widens (\edfig~\ref{fig:emislead}j).
They also degrade it, by 1--3\%, in the Middle East, where dust dominates as well\cite{mcduffie2021source}.
Second, where sources are intense every day, the start-day field already reflects them.
In the North China Plain and the Indo-Gangetic Plain, where residential, industrial and energy emissions are large\cite{mcduffie2021source}, the gain is small or absent, with 95\% intervals that include zero (Methods).
A monthly inventory adds little to such a start day, and inventories are also uncertain there\cite{saikawa2017india,yang2025china}.
In South Asia, the emissions even degrade the ensemble mean slightly in the first 3 days and have no effect thereafter.

On 16 May 2022, \pmtf{} around Kano, Nigeria, rose sharply in the \nrev{estimates}.
The forecasts started on the three preceding days traced the arriving air to the dust over Niger and Chad, hundreds of kilometres to the north-east (Fig.~\ref{fig:transport}a--d).
In the forecasts issued 1 and 2 days before, Kano had a probability above one half of exceeding 100\,\textmu g\,m$^{-3}$.
In the forecast issued 3 days before, most members kept the plume north of the city (Fig.~\ref{fig:transport}e--g).

\FloatBarrier
\subsection*{\nrev{Wildfire smoke}}

Wildfire smoke harms people and wildlife far from the flames\cite{sanderfoot2021review}, so a useful forecast must place it at the scale of the towns and nature reserves that it reaches.
Forecasting smoke days ahead is difficult for two reasons.
First, smoke from a fire that ignites after a forecast is issued cannot be foreseen by any system.
Second, Foehn knows the fires only through their emissions on the start day, estimated from satellite observations\cite{kaiser2012gfas}, so it cannot know whether they will grow or die down.
In this case, its 16 members give at every place the probability that \pmtf{} exceeds 25\,\textmu g\,m$^{-3}$, the WHO interim target 4.

In July 2022, during the second-worst fire season of the European Union by burnt area, two fires that started on 12 July burnt about 18,000 hectares of pine forest in the Gironde, in southwestern France, before being contained\cite{sanmiguel2023fires}.
One of them burnt beside the R\'eserve Ornithologique du Teich, a bird reserve on the Arcachon basin, where the smoke raised \pmtf{} to its peak on 18 July, the only day from 13 to 26 July above 25\,\textmu g\,m$^{-3}$ in the estimates (Fig.~\ref{fig:smoke}a,f).
Issued on 11 July, before the fires ignited, Foehn's forecast for 18 July gives almost no chance of exceeding this threshold near the fires, as any forecast issued before an ignition must (Fig.~\ref{fig:smoke}b).
Issued on 13 July, 5 days ahead, it gives a probability of at least one in four to 94\% of the area in which the estimates put \pmtf{} above 25\,\textmu g\,m$^{-3}$, with a mean probability of 0.48 inside this area and 0.12 outside it (Fig.~\ref{fig:smoke}d).
It gives the reserve a probability of approximately one third (Fig.~\ref{fig:smoke}g).
However, its zone of likely exceedance is five times larger than in the estimates, mainly over the forest of the Landes.

The warning at the reserve held across lead times.
For 18 July, the members rose towards the estimated \pmtf{} as the fires entered the inputs: one of 16 exceeded 25\,\textmu g\,m$^{-3}$ 7 days ahead, and five did so from 5 to 3 days ahead (Fig.~\ref{fig:smoke}h).
This is the only target day to keep a probability of 0.19--0.31 at every lead time from 1 to 5 days. On the other days, the probability stays mostly below 0.19, with isolated false alarms of up to 0.44 (24 July, 6 days ahead; Fig.~\ref{fig:smoke}g).
This case shows that, once fires appear in its inputs, Foehn can give the probability that smoke pushes \pmtf{} above a health threshold at the scale of a nature reserve 5 days ahead.


\FloatBarrier
\subsection*{Discussion}
We introduce Foehn, a generative model that forecasts daily fine particulate matter (\pmtf{}) at kilometre-scale resolution across all inhabited land, up to a week ahead. By combining a transport model with flow matching, Foehn produces physically consistent forecasts with calibrated ensembles. Its forecasts outperform both operational forecasts and other neural networks, while detecting local extremes days earlier. Three limitations remain. First, Foehn cannot yet deliver real-time forecasts at kilometre-scale resolution because its initial estimates are generated retrospectively. Real-time deployment would require a method to derive kilometre-scale \pmtf{} estimates from coarser, routinely available forecasts. A more practical near-term approach is to deploy Foehn in cities that already provide street-scale \pmtf{} maps based on live measurements, such as London \cite{stidworthy2018london,cercairtext}. Second, errors and regional biases in the high-resolution \pmtf{} estimates can propagate through Foehn or be amplified. Although we quantitatively evaluate forecasts against observations in countries with live monitoring networks, forecast reliability remains uncertain in regions without comparable measurements. Third, all ensemble members rely on the same deterministic weather forecast, while the emission inputs omit desert dust and misrepresent the locations of other sources. Regional emission inventories \cite{zheng2018trends}, together with daily fire and dust emission estimates, could help address these limitations.

In conclusion, combining explicitly modelled physics with learned dynamics offers a path towards accurate and interpretable air-quality forecasting. This approach could extend to other airborne quantities, including individual components of fine particulate matter, coarser particles, and trace gases. Kilometre-scale ensemble forecasts could enable neighbourhood-level warnings, exposure assessments in regions with sparse monitoring, and better protection of nature reserves from smoke. Despite its current limitations, Foehn demonstrates the potential for air-quality forecasts at the neighbourhood scale, bringing actionable warnings within reach of millions of people who remain invisible to coarser-resolution maps.


\OrigFloatBarrier   
\section*{Methods}

\subsection*{Problem statement}

Let $x_d$ be the field of daily-mean surface \pmtf{} on day $d$, on a regular latitude--longitude grid of 0.01$^\circ$
($\approx$1\,km$\times$1\,km). A forecast is issued at the end of a start day $d$ for a target day $d+h$, where $h\in\{1,\dots,7\}$ is the
lead time in days. \model{} is a generative model: it draws samples from the conditional distribution
\begin{equation}
p\big(x_{d+h}\mid c_{d,h},m,y,e,w\big),
\label{eq:problem}
\end{equation}
where $c_{d,h}$ is the 1\,km \pmtf{} of days $d$ and $d-1$; $m$ the static maps (elevation and a land mask); $y$ the
coarse-resolution (0.25$^\circ$) \pmtf{} of day $d$; $e$ the weather and wildfire emissions of day $d$ and the forecast weather of
day $d+h$; and $w$ the transport of the air between days $d$ and $d+h$. \nrev{Each sample is a member of the forecast, and $M$ members form an ensemble.} Every lead time is forecast directly from
day $d$, so that errors cannot accumulate through an autoregressive rollout. The forecasts are issued as if the 1\,km \pmtf{} of
day $d$ were available at its end. These data are produced later, so every 1\,km forecast in this study is retrospective.

\subsection*{Datasets}

The training data is the daily 1\,km \pmtf{} of the Global High Air Pollutants dataset (GHAP, version 1)\cite{wei2023global} for
2017--2022, \nrev{gap-free estimates} derived from satellite data and meteorological reanalyses and fitted to station measurements (we refer to it as the \nrev{1\,km estimates} in the main text). The years 2017--2020 train the models, 2021 validates
them and 2022 is the test year.

\model{} does not forecast the weather: it takes it from the operational high-resolution forecast (HRES) of the European Centre
for Medium-Range Weather Forecasts (ECMWF), run at 00\,UTC on the start day $d$. This single forecast provides both the weather of each target day in $e$ and \nrev{the weather that drives the transport} $w$. Only
the weather of the start day comes from the ERA5 reanalysis\cite{hersbach2020era5}\nrev{, released about 5 days after real time, which an operational analysis would replace in real time}.

Wildfire emissions come from the Global Fire Assimilation System (GFAS)\cite{kaiser2012gfas} of the Copernicus Atmosphere
Monitoring Service (CAMS), for day $d$ only; monthly anthropogenic emissions come from CAMS-GLOB-ANT v6.2\cite{soulie2024cams}, a
global inventory extended to recent years by extrapolation. The residents of each 1\,km pixel come from GHS-POP
R2023A\cite{schiavina2023ghspop}. Since the 1\,km estimates inherit bias and pattern errors, we further evaluate the forecasts at 6,324 stations of six official networks: the European Environment
Agency (EEA; 2,029 stations), the China National Environmental Monitoring Centre (CNEMC\cite{cnemc}; 1,338), the Air Quality
System of the US Environmental Protection Agency (EPA; 1,016), AirKorea (619), the Japanese national network (949) and the Central
Pollution Control Board of India (CPCB; 373). \nrev{Daily means are over UTC days, like those of GHAP\cite{wei2023global} and of every forecast, from at least 12 hourly values (European network: official daily means, offset by 1\,h; US sites without hourly values are left out), and stations with at least 30 valid days are kept.}
The station measurements are never used to train or select \model{}.

\subsection*{The \model{} model}

\paragraph{Transport-informed latent field: FLEXGREEN.} The concentration at a receptor is the sum of the earlier field and of the
sources met by its air, each weighted by the Green's function of the transport, the sensitivity of the concentration at the
receptor to the air at each earlier place and time\cite{seibert2004source}. A backward Lagrangian particle model estimates this
function by releasing particles of air at the receptor and following them back in time. FLEXGREEN is our re-implementation, on
graphics processors, of the backward mode of FLEXPART\cite{stohl2005flexpart,pisso2019flexpart}, driven by the HRES \nrev{weather}.
\nrev{It follows FLEXPART's published equations: Petterssen advection\cite{petterssen1940weather} with 15-min time steps; the height of the mixed layer, the turbulent layer next to the surface, from the bulk Richardson number\cite{vogelezang1996evaluation}; Hanna turbulence\cite{hanna1982applications} with a random-displacement model\cite{thomson1987criteria}; and constant diffusivities above the mixed layer.}
\nrev{It differs from FLEXPART in three ways.}
\nrev{(i) The weather comes from the 6-hourly HRES fields on eight pressure levels, averaged over 1$^\circ$ boxes, and, without surface fluxes, the friction velocity, Obukhov length and convective velocity are derived from Businger--Dyer profiles\cite{businger1971flux}, as in HYSPLIT\cite{draxler1998hysplit}.}
\nrev{(ii) Where turbulent mixing is faster than one time step, the heights of particles in the mixed layer are redrawn uniformly between the ground and its top, the stationary state of a well-mixed layer.}
\nrev{(iii) Removal takes FLEXPART's simplest forms: a dry deposition velocity of 0.05\,cm\,s$^{-1}$ in the lowest 30\,m, and wet scavenging $\Lambda=AI^B$ below precipitating clouds, with $I$ the precipitation rate in mm\,h$^{-1}$, $A=10^{-4}$\,s$^{-1}$ and $B=0.8$.}
 For
every 0.5$^\circ$ land cell (a receptor $r$), start day $d$ and lead time $h$, $P=32$ particles are released \nrev{at random times} during the target day
between 0 and 100\,m above the ground and followed back $h$ days. Each particle $p$ stores its origin $\mathbf x_p$ on day $d$; its
height there divided by the maximum height of the mixed layer on that day, $\eta_p$ (the
air was in the mixed layer when $\eta_p\le1$); and its survival $s_p\in[0,1]$, the fraction of its aerosol not removed by
deposition and rain on the way. \nrev{Along the paths, the GFAS emissions (three fire terms) of day $d$, spread from the ground to the larger of the injection height and the mixed-layer height. The residence time below 100\,m\cite{seibert2004source}, weighted by survival, times the monthly CAMS-GLOB-ANT emission fluxes, including black and organic carbon, SO$_2$, NO$_x$, NH$_3$ and CO.}

\paragraph{Transport-informed latent field: tracers.} Transport is linear, so it commutes with any linear combination of transported quantities. An encoder
of two $3\times3$ convolutions turns the 0.25$^\circ$ fields of day $d$ (the standardized $y$, its validity mask and the logarithms of the three GFAS variables: the fire emission of particulate matter, the fire radiative power
and the injection height) into 32 latent tracers $f=(f_1,\dots,f_{32})$ at 0.5$^\circ$, maps whose content is
learned. Each receptor $r$ gathers them at the origins of its particles:
\begin{equation}
g_r=\frac{\sum_{p}\omega_p\,f(\mathbf x_p)}{\sum_p \omega_p},\qquad \omega_p=s_p\,\mathbf{1}[\eta_p\le1],
\label{eq:gather}
\end{equation}
where the sums run over the particles released at $r$ \nrev{and} $f(\mathbf x_p)$ is interpolated bilinearly, so that only air that was in the mixed layer on day $d$ is read. The operator has no
parameters and is differentiable in $f$, so the encoder learns what to transport from the forecast error. With the advected
\pmtf{}, the share $a_r$ of the air that came from above the mixed layer, the mean survival and the wildfire and anthropogenic
emission terms, each receptor receives 44 transport channels, which represent $w$.

\paragraph{Generator network.} The generator $v_\theta$, a neural network with parameters $\theta$, predicts the velocity of the
flow at 1\,km over a window of $4{,}192\times6{,}992$ pixels in one pass. It is a
U-Net\cite{ronneberger2015unet} of residual blocks from 2 to 32\,km that receives $x_\tau$, $c_{d,h}$ and $m$ at 1\,km and $y$, $e$
and the transport channels at 16\,km. At 64\,km, four transformer blocks\cite{vaswani2017attention} apply global self-attention
across the window, so that information crosses the continent in one layer.
\nrev{Sinusoidal embeddings of the flow time and of the lead time modulate every block\cite{perez2018film}.}
The whole model has 14.2 million parameters.

\paragraph{Global forecasts.} Twenty-one overlapping windows cover the land between 84$^\circ$N and 60$^\circ$S and are blended \nrev{with weights rising as $\sin^2$ over 600\,km from each edge}. Each member uses one global noise field, which every window cuts at its own position, so that overlapping
windows integrate the same noise.

\subsection*{Training methods}

Training has three stages: base training by flow matching, fine-tuning with the anthropogenic emissions, and calibration
fine-tuning on the continuous ranked probability score (CRPS; see \textit{Evaluation and verification metrics}). An epoch is one
pass over the 9,730 training pairs of start day and lead time of 2017--2020.

\paragraph{Flow matching.} We learn a velocity field $v_\theta$ that carries Gaussian noise to the distribution of
equation~\eqref{eq:problem}\cite{lipman2023flow,liu2023flow,albergo2023building}. Let $x_1=(x_{d+h}-15)/20$ be the standardized
target (concentrations in \textmu g\,m$^{-3}$), $x_0$ a field of independent standard Gaussian noise and
$x_\tau=(1-\tau)x_0+\tau x_1$ the straight path between them, where $\tau\in[0,1]$ is the flow time. The network learns the
velocity $x_1-x_0$ along this path:
\begin{equation}
\begin{split}\mathcal L(\theta)=\mathbb E\Big[\frac{1}{|\Omega|}\sum_{i\in\Omega}\big(&v_\theta(x_\tau,\tau,h;\,c_{d,h},m,y,e,w)_i\\&-(x_1-x_0)_i\big)^2\Big],\end{split}
\label{eq:fm}
\end{equation}
where $\Omega$ is the set of valid land pixels of the window and $i$ indexes its pixels. A member is drawn by integrating
$\mathrm dx/\mathrm d\tau=v_\theta$ from noise at $\tau=0$ to a forecast at $\tau=1$ with four Euler steps; each noise field gives
one member. For half of the training samples, the noise level $\sigma=1-\tau$, drawn uniformly on $[0,1]$, is mapped to
$\kappa\sigma/(1+(\kappa-1)\sigma)$ with $\kappa=25$, which moves training towards high noise
levels\cite{hoogeboom2023simple,esser2024scaling}; the four Euler steps are placed at the quantiles of this distribution of $\tau$.

\paragraph{Base training and emission fine-tuning.} The base model is trained for 60 epochs on 32 graphics compute dies (GCDs;
19\,h), with the HRES weather and transport of the training years. It is optimized with the AdamW optimizer\cite{loshchilov2019decoupled} and a one-cycle learning-rate
schedule\cite{smith2019super}. It uses the 38 transport channels without the six anthropogenic emission terms. It is then fine-tuned for 12
epochs, with the same loss, with the six additional anthropogenic emission terms appended. Their input weights start at zero, so that the
uncertain inventories cannot destabilize the base model. The checkpoint after its third epoch, selected on 2021, is the model with emissions.
The ablation without transport is trained like the base model, with the transport channels set to zero.

\paragraph{Calibration fine-tuning.} Trained by flow matching alone, the ensembles were too narrow: on 2021 the spread--skill
ratio of four members was 0.51 one day ahead and 0.43 after a week. We therefore fine-tuned the model on the CRPS, as recent
weather models do\cite{lang2024aifs}: each training sample draws two members exactly as a forecast does, and the loss is their
fair CRPS against GHAP\cite{ferro2014fair} plus the flow-matching loss, with the gradient flowing through the sampler. After two
epochs, the spread--skill ratio on 2021 was 0.92 one day ahead and 0.82 after a week, and the CRPS was 12\% lower one day ahead,
for a 2\% increase in the error of the ensemble mean (\edfig~\ref{fig:fair}h); on 2022 the ratio was 0.96--0.99 at every lead time. \nrev{In total, training the final model took 25.8\,h on 32 GCDs.}

\paragraph{Design, selection and test.} \nrev{The European benchmark (see \textit{Benchmarks}) set the design principles, and the configuration of the global model was fixed before training. Its ablations did not change it:} on 2021, the model trained with ERA5 weather had a 1\% lower CRPS summed over the
seven lead times (0.6\% higher one day ahead), and the planned configuration was kept.

\subsection*{Benchmarks}

\paragraph{European benchmark.} A controlled benchmark over Europe is used for ablation studies of the design: five
spatiotemporal networks, ConvLSTM\cite{shi2015convolutional}, SimVP\cite{gao2022simvp}, Earthformer\cite{gao2022earthformer},
ClimaX\cite{nguyen2023climax} and TopoFlow\cite{kheder2026topoflow}, were retrained with exactly the inputs of the flow-matching
network and scored on 2022 against GHAP and 2,971 EEA stations. \nrev{The flow-matching network scored better than each other system in 319 of 324 comparisons of score, lead time and set of days, and was within 2\% in the five others} (Extended Data Figs.~\ref{fig:eu_lead} and \ref{fig:eu_extremes}). \nrev{Networks} trained at 1\,km
and applied tile by tile left blocky artefacts at the tile boundaries (\edfig~\ref{fig:km}).

\paragraph{Comparison in Aurora's protocol.} \nrev{To compare \model{} with Aurora on its own terms, we retrained \model{} in the protocol of Aurora's air-pollution model}\cite{bodnar2025aurora} (Foehn-CAMS; Extended Data
Figs.~\ref{fig:auroraset} and \ref{fig:aurorareg}): on the 0.4$^\circ$ grid of the CAMS analysis, from the analyses of 00\,UTC and
of 12\,h earlier, it forecasts the instantaneous \pmtf{} every 12\,h up to 120\,h. Foehn-CAMS keeps \model{}'s components
(FLEXGREEN, 32 transported tracers, the HRES forecast of the initial day as weather, and the same network and loss), applied to the
whole globe at once, with 9.9 million parameters, 131 times fewer than Aurora. All its inputs are available in real time. It was
trained once and its final checkpoint scored once, on forecasts initialized at 00\,UTC from May to November 2022, against the CAMS
analysis.

\subsection*{Evaluation and verification metrics}

All scores are on the forecasts issued every second start day of 2022, one to seven days ahead. \model{}'s forecasts have four members ($M=4$) unless stated otherwise and are read at the 1\,km pixel of
each station. The CAMS global forecast\cite{peuch2022cams} and Aurora (0.4$^\circ$, one to four days ahead) are interpolated
bilinearly. Because \model{} uses the whole start day, it could be issued at the end of day $d$, 24\,h after the CAMS run of that
date. Comparisons at equal issuance time therefore use the CAMS and Aurora forecasts of 00\,UTC on day $d+1$. The climatology of a
place and calendar day is the mean GHAP of 2017--2020 within $\pm$7 days of that day. Systems are compared on identical
station-days or, against GHAP, on identical 0.25$^\circ$ land cells and days. Read at the station pixel on the same day, GHAP
itself has an RMSE against the measurements of 2022 of \nrev{2.4\,\textmu g\,m$^{-3}$ at the Chinese stations, 2.6 at the US, 6.6 at the Japanese, 8.5 at the Korean, 11.2 at the European and 36.5 at the Indian stations (bias $-$5.9), and 11.0 over the six networks}.

We report the root-mean-square error (RMSE) and the bias of the ensemble mean. For uncertainty analysis, we use the CRPS, a strictly proper
score\cite{gneiting2007strictly} that rewards both accuracy and calibration, estimated without ensemble-size bias\cite{ferro2014fair}
and equal to the absolute error for a deterministic forecast; and the spread--skill ratio\cite{fortin2014why}, which is 1 for a
calibrated ensemble and below 1 for an ensemble that is too narrow. Local extremes are the days above the local 90th, 95th or 99th
percentile of 2017--2020\nrev{, computed for each 0.25$^\circ$ cell and calendar day ($\pm$15 days) from the GHAP block means} (the moderate, severe and extreme tiers, labelled against the local climate as in
TailWeather\cite{liu2026tailweather}); their detection is scored by the symmetric extremal dependence index
(SEDI)\cite{ferro2011extremal}, which does not depend on how rare the events are. The relative economic value\cite{richardson2000skill} of warnings is the
fraction of the saving of a perfect forecast that a user who protects at a cost $C$ against a loss $L$ obtains, as a function of
$C/L$; for \model{} it is maximized over probability thresholds on the 2022 data, a retrospective potential value. Differences
between systems are tested with a paired moving-block bootstrap over start dates\cite{kunsch1989jackknife} (blocks of seven days,
10,000 resamples, 95\% intervals).

Relative gains are computed from unrounded scores: at a lead time of 1 day at the stations, the RMSE of \model{} is 12.68\,\textmu g\,m$^{-3}$, against 15.23 for the CAMS forecast and 15.72 for Aurora (reductions of 16.7\% and 19.3\%).
\nrev{At the 2,029 European stations, on the same station-days, the climatology of GHAP at the station pixel has an RMSE of 11.0\,\textmu g\,m$^{-3}$, against 9.8 for \model{} 1 day ahead, 10.7 for the CAMS forecast and 10.8 for Aurora.}

At the stations, with the thresholds used over land (the local 99th percentile of GHAP in 2017--2020 at the station pixel), the SEDI of the ensemble mean of \model{} is 0.45 at a lead time of 1 day and 0.34 at 4 days, against 0.33 and 0.25 for the CAMS forecast and 0.34 and 0.26 for Aurora, and the same holds at the 95th percentile.
With each station's own 99th percentile of 2022, the three systems are equal at 1 day (0.44--0.45), and the CAMS forecast and Aurora score higher from 2 days onwards (0.45 and 0.48, against 0.35).
The measurement of the start day at the station, which exists only where there is a station, scores higher than every forecast with the thresholds of GHAP (0.73 at 1 day and 0.47 at 4 days).

\subsection*{Diagnostics and case studies}

A resident is hidden by the 40\,km grid when the \pmtf{} of their 1\,km pixel exceeds a threshold and the mean of their 40\,km
cell, a $40\times40$-pixel block like the 0.4$^\circ$ cells of CAMS, does not. The value of the proposed particle transport is the reduction of the CRPS against GHAP of the base model relative to the same network trained without the transport channels. The value of the emissions is the reduction of the RMSE of the ensemble mean and of the CRPS of the model with emissions relative to the base model. In the three regions not shown in \edfig~\ref{fig:emislead},
the gain of the emissions in the RMSE of the ensemble mean is 7--11\% in Oceania and 3--5\% in Southeast Asia from one to seven
days ahead, and $-3$ to $-1$\% in the Middle East. In the North China Plain (34--41$^\circ$N, 112--120$^\circ$E) and the
Indo-Gangetic Plain (24--31$^\circ$N, 74--88$^\circ$E), it is between $-0.1$ and 3.2\% and within $\pm$1.5\%, respectively, with
95\% intervals that include zero at every lead time. \nrev{After a week, the spread--skill ratio of the model with emissions is 0.51, against 0.43 for the base model; it is within 0.02 of the base model in Europe, North America and East Asia but 0.12 higher in Africa, so part of the gain in CRPS at long lead times comes from a wider ensemble in Africa, where the emissions do not improve the ensemble mean.}

Each case was chosen by a rule fixed before any forecast of it was examined. The global day (Fig.~\ref{fig:overview}) and the
exposure case (Fig.~\ref{fig:hidden}) are the day of 2022 with the highest population-weighted GHAP (21 January). The
extreme-event case (Fig.~\ref{fig:realism}) is the day of that January with the highest mean at the official stations of
Beijing--Tianjin--Hebei (25 January); the four basins of Fig.~\ref{fig:overview}b are each monitored by a different official
network. Among twenty large cities fixed in advance, the other exposure cases (\edfig~\ref{fig:exposure}) are the city-days with
the most residents above 75\,\textmu g\,m$^{-3}$ in 40\,km cells below it (Cairo, 2 January; Santiago de Chile, 20 June), and the
transport case (Fig.~\ref{fig:transport}) is the African arrival day with the largest rise of \pmtf{} carried by the air (Kano,
16 May). The smoke case (Fig.~\ref{fig:smoke}) concerns the largest fires of 2022 in France, and its day, 18 July, has the highest GHAP
of 13--26 July around the R\'eserve Ornithologique du Teich.

\subsection*{Use of generative AI}
 \nrev{An LLM-based coding assistant was used to write analysis and figure code, to restructure and draft the manuscript and to cross-check its numbers against the result files. The authors reviewed all AI-assisted text, code and claims against the code and result files and take full responsibility for the content of this work.}

\subsection*{Data availability}
 All data used are public. GHAP daily 1\,km \pmtf{} (version 1) is available from Zenodo
(Wei et al.\cite{wei2023global}); ERA5, the CAMS global forecasts and analyses, CAMS-GLOB-ANT and GFAS from the Copernicus
Climate and Atmosphere Data Stores; the IFS HRES forecasts from WeatherBench~2\cite{rasp2024weatherbench2}; GHS-POP R2023A from the Joint Research Centre; elevation from the ETOPO 2022 global relief model of NOAA (30 arc-second
ice-surface grid)\cite{etopo2022,macferrin2025etopo}; and the station measurements from the European Environment Agency, the China National
Environmental Monitoring Centre\cite{cnemc}, the US Environmental Protection Agency (Air Quality System), AirKorea, the National
Institute for Environmental Studies of Japan and the Central Pollution Control Board of India, partly through
OpenAQ\cite{openaq}.

\subsection*{Code availability}
 The code of FLEXGREEN, of the training and inference of \model{} and of every analysis and
figure is available at \url{https://github.com/AmmarKheder/Foehn} under the MIT licence, and the version used here (v1.0.0)
is archived at Zenodo (\url{https://doi.org/10.5281/zenodo.23337751}) and in the Software Heritage archive (SWHID
\url{swh:1:rev:b523bc369b9a68e730ee5a29caf002ba2d83c422;origin=https://github.com/AmmarKheder/Foehn}). The trained weights of the five models evaluated here, with
their checksums and the inference commands, are archived at Zenodo (\url{https://doi.org/10.5281/zenodo.23337591}); access to the
files is restricted to the editors and referees during peer review and will be opened upon publication.


\begingroup\small
\paragraph{Acknowledgements} This work was supported by the Finnish Ministry of Education and Culture's Pilot for Doctoral
Programmes (Pilot project Mathematics of Sensing, Imaging and Modelling).
We acknowledge the ACCC Flagship, funded by the Academy of Finland (grant no. 337549). We acknowledge CSC -- IT Center for
Science, Finland (\url{https://csc.fi}), for awarding this project generous access to the LUMI supercomputer, owned by the
EuroHPC Joint Undertaking, hosted by CSC (Finland) and the LUMI consortium. Thanks to the city of Lahti (Finland) for financial
support. We thank the providers of the data listed in Data availability.
\paragraph{Author contributions} A.K. conceived and designed the study, developed Foehn and the FLEXGREEN transport code, ran
the experiments, analysed the results and wrote the paper. Z.-S.L. supervised the machine learning development and evaluation, contributed to
methodology development and manuscript writing. M.B. supervised the atmospheric science aspects, provided overall project
direction, and secured funding. All authors reviewed and edited the manuscript.
\paragraph{Competing interests} The authors declare no competing interests.
\paragraph{Correspondence} Correspondence and requests for materials should be addressed to Ammar Kheder
(\href{mailto:ammar.kheder@lut.fi}{ammar.kheder@lut.fi}).
\endgroup

\titleformat{\section}{\sffamily\bfseries\large\color{accent}}{}{0pt}{}[{\color{accent}\titlerule[0.6pt]}]
\bibliographystyle{unsrtnat}
\bibliography{refs}

\onecolumn
\titleformat{\section}{\sffamily\bfseries\Large\color{accent}}{\thesection}{0.6em}{}[{\color{accent}\titlerule[0.8pt]}]
\section*{Extended Data}
\setcounter{table}{0}\renewcommand{\tablename}{Extended Data Table}
\setcounter{figure}{0}\renewcommand{\figurename}{Extended Data Fig.}
\begin{figure}[!ht]
\centering\includegraphics[width=\textwidth]{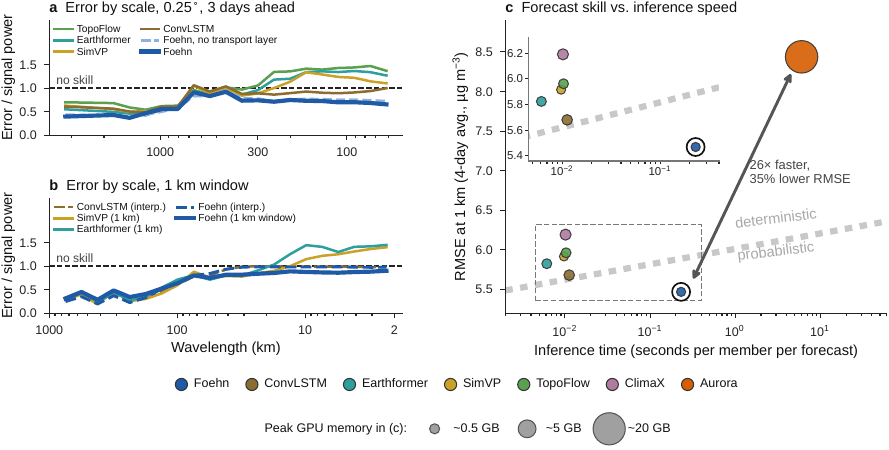}
\caption{\textbf{Over Europe, Foehn is 26 times faster than Aurora, with a 35\% lower error at 1\,km.} \textbf{a}, Power of the forecast error
relative to the power of the GHAP anomaly, by wavelength, at 0.25$^\circ$ over central Europe (61 start days of 2022, three
days ahead): below one, skill at that wavelength; at one, no skill; above one, structure in the wrong place. \textbf{b}, The
same at 1\,km on the regional window from Paris to Florence, for the models trained natively at 1\,km and for the continental
0.25$^\circ$ models interpolated to 1\,km. \textbf{c}, RMSE at 1\,km (mean over leads one to four days, Europe 2022) against
the time to produce one member of one forecast on one MI250X compute die; the marker area is the peak GPU memory. Foehn is
26 times faster than Aurora with 35\% lower error; the inset enlarges the boxed region.}
\label{fig:speccost}
\end{figure}
\clearpage

\begin{figure}[!ht]
\centering\includegraphics[width=\textwidth]{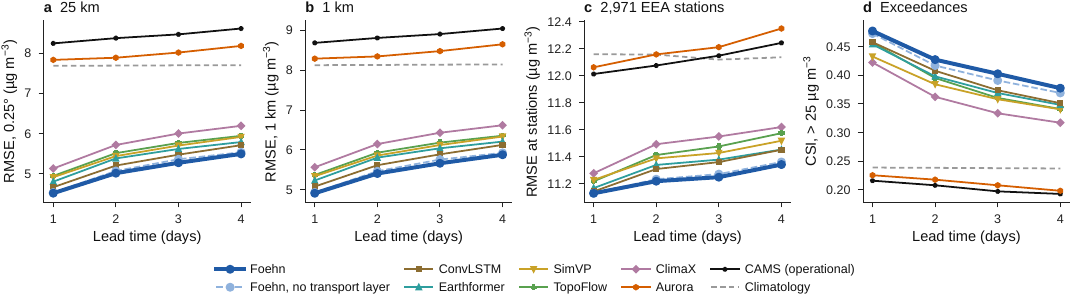}
\caption{\textbf{Over Europe, Foehn's network is more accurate than five networks retrained with identical inputs, the CAMS forecast and Aurora.} Root-mean-square error of the forecasts of
2022 against the 0.25$^\circ$ block means (\textbf{a}), the 1\,km pixels (\textbf{b}) and 2,971 European Environment Agency
stations (\textbf{c}), and critical success index of exceedances of 25\,\textmu g\,m$^{-3}$ (\textbf{d}), versus lead time:
Foehn's flow-matching network (its 0.25$^\circ$ European version) with and without the learned transport layer, ConvLSTM, Earthformer, SimVP, TopoFlow and
ClimaX retrained with identical inputs, the CAMS operational forecast, Aurora and climatology. Means over three seeds.}
\label{fig:eu_lead}
\end{figure}
\clearpage

\begin{figure}[!ht]
\centering\includegraphics[width=\textwidth]{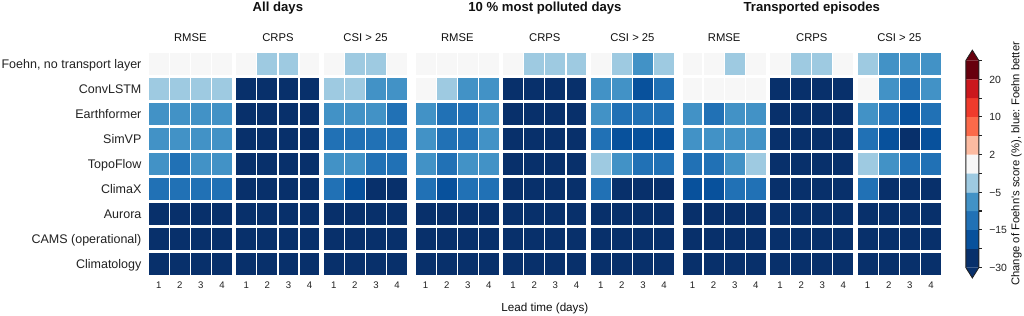}
\caption{\textbf{Over Europe, Foehn scores better than each other system in 319 of 324 comparisons, on all days, polluted days and transported episodes.} Scorecard of the
change of Foehn's score relative to each system (rows), for the root-mean-square error (RMSE), the continuous ranked probability
score (CRPS; absolute error for point forecasts) and the critical success index of exceedances of 25\,\textmu g\,m$^{-3}$ (CSI, sign
reversed so that blue always means Foehn better), one to four days ahead, over Europe in 2022 at 0.25$^\circ$: on all days (left),
on the 10\% of days with the largest polluted area (centre) and during the four documented transported episodes (right). Means over
three seeds; Foehn is better in 319 of the 324 cells, and the five others are within 2\%. Systems as in Extended Data
Fig.~\ref{fig:eu_lead}.}
\label{fig:eu_extremes}
\end{figure}
\clearpage
\begin{figure}[!ht]
\centering\includegraphics[width=\textwidth]{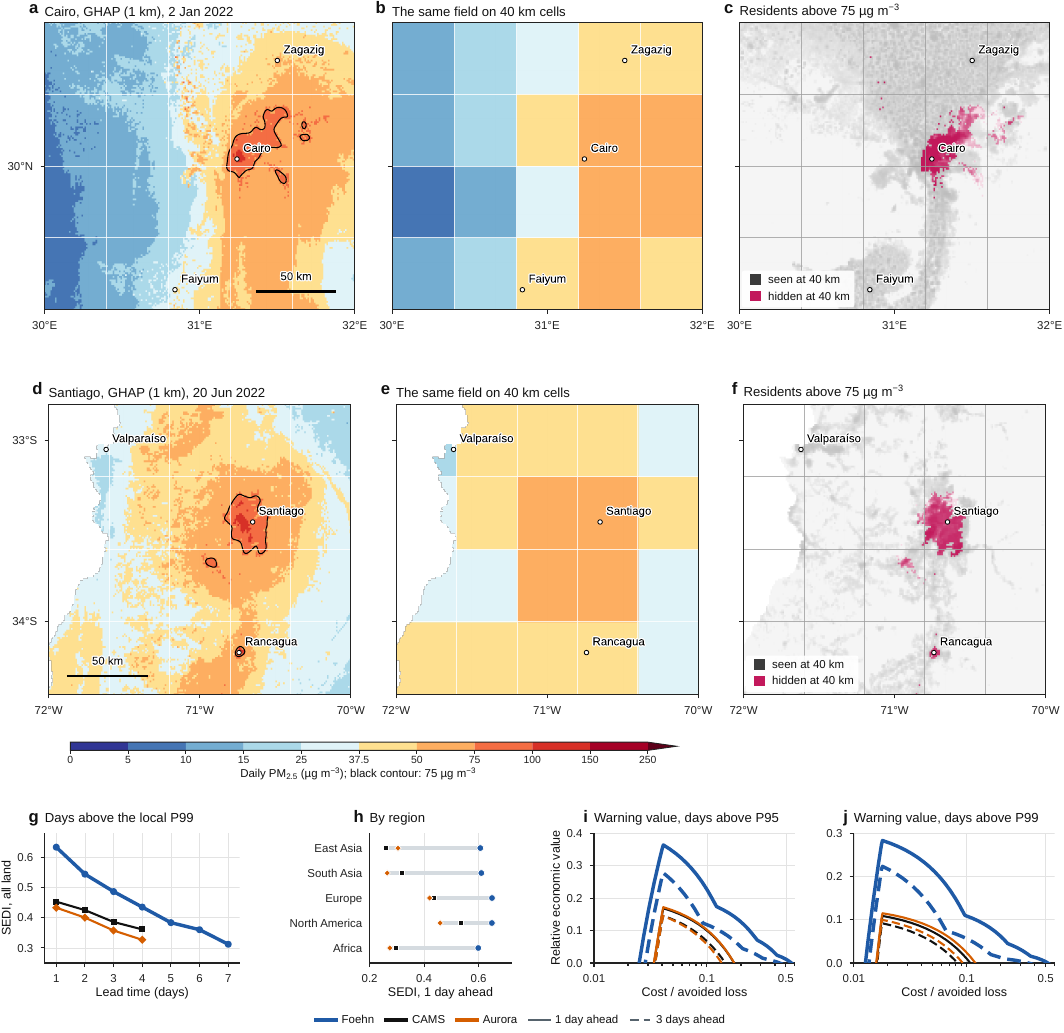}
\caption{\textbf{\nrev{Coarse grids hide whole cities from warnings, and Foehn's kilometre-scale forecasts detect local extremes earlier and with more value.}} \textbf{a}--\textbf{c}, Cairo on 2 January 2022 and \textbf{d}--\textbf{f}, Santiago de Chile on 20 June 2022, the largest
hidden city-days of Africa and of the Americas among twenty large cities fixed in advance (Methods), as in
Fig.~\ref{fig:hidden}b--d: GHAP at 1\,km with the contour of 75\,\textmu g\,m$^{-3}$, the same field on 40\,km cells, and the
residents above 75\,\textmu g\,m$^{-3}$ seen (grey) or hidden (pink) by the 40\,km cells. Around Cairo, 16.6 million residents were
above the threshold and every one of them hidden; around Santiago, 5.8 million, all hidden, the basin of the city averaged with
the surrounding slopes of the Andes and the coastal range. \nrev{\textbf{g}, \textbf{h}, Detection of local extremes over land in 2022, the days above the local 99th
percentile of 2017--2020: symmetric extremal dependence index (SEDI) by lead time (\textbf{g}; the CAMS and Aurora forecasts cover
4 days) and by region 1 day ahead (\textbf{h}), for Foehn's ensemble mean and the CAMS and Aurora\cite{bodnar2025aurora} forecasts
issued at the same time. \textbf{i}, \textbf{j}, Retrospective potential relative economic value of warnings of days above the
local 95th (\textbf{i}) and 99th (\textbf{j}) percentiles at the monitoring stations, against the ratio of the cost of protection to
the loss avoided, 1 day (solid) and 3 days ahead (dashed).}}
\label{fig:exposure}
\end{figure}
\clearpage

\begin{figure}[!ht]
\centering\includegraphics[width=\textwidth]{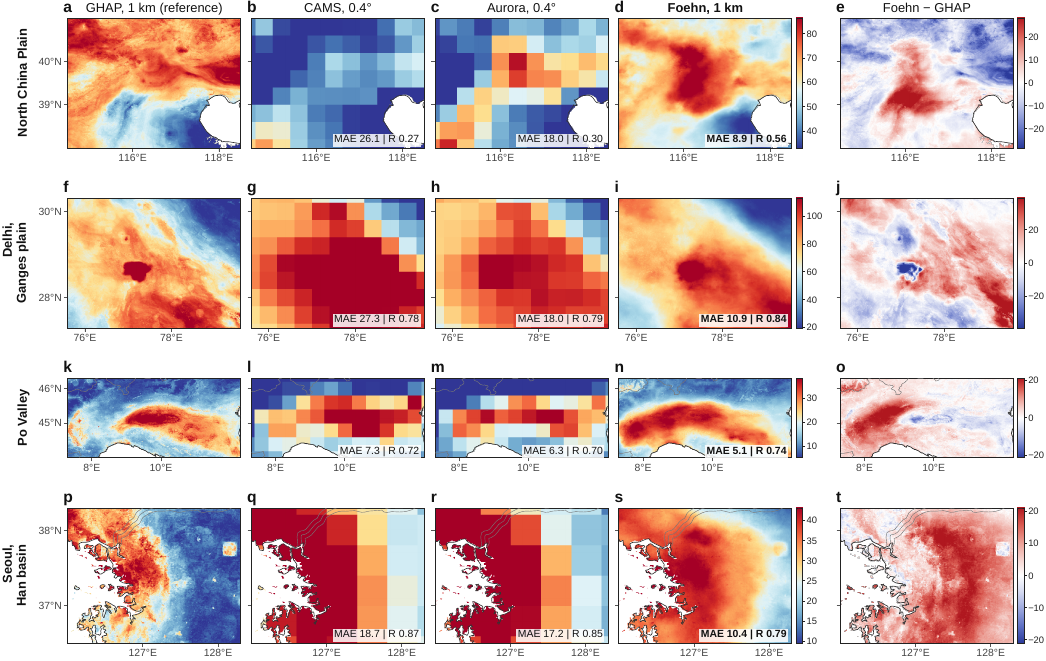}
\caption{\textbf{On 21 January 2022, Foehn resolves at 1\,km the structure of four polluted regions that the 0.4$^\circ$ forecasts blur.} Daily mean \pmtf{} on 21 January 2022 over the North China Plain,
the Indo-Gangetic Plain around Delhi, the Po Valley, and Seoul and the Han basin (regions chosen
before the comparison): \nrev{the 1\,km estimates} (first column), the CAMS forecast and Aurora issued at 00\,UTC on 21 January
(0.4$^\circ$), Foehn's ensemble mean issued at the end of 20 January (1\,km), and Foehn minus the reference. Each row has its own
colour range (2nd--98th percentiles of the reference). Numbers: mean absolute error (MAE, \textmu g\,m$^{-3}$) and spatial
correlation against the reference over the land pixels of the panel, the 0.4$^\circ$ forecasts read at the 1\,km pixels.}
\label{fig:regions}
\end{figure}
\clearpage

\begin{figure}[!ht]
\centering\includegraphics[width=\textwidth]{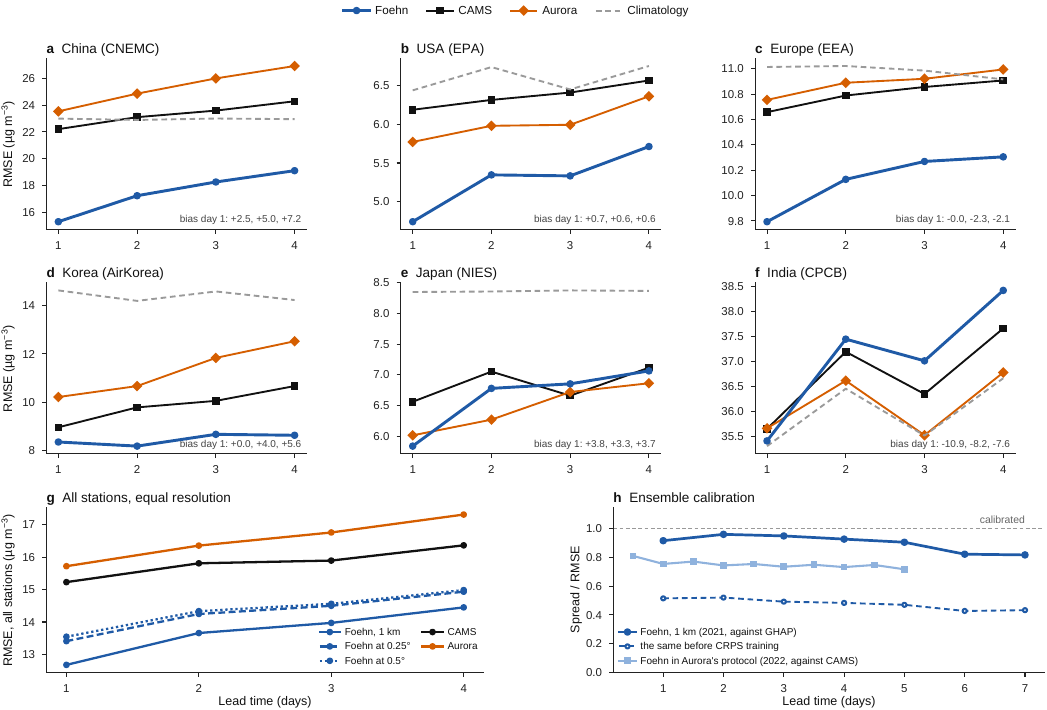}
\caption{\textbf{At the stations of six official networks, Foehn's error 4 days ahead is 12\% lower than that of the CAMS forecast and 17\% lower than Aurora's.} \textbf{a}--\textbf{f}, Root-mean-square error (RMSE) of the daily
\pmtf{} forecasts of 2022 by lead time at the stations of each official network, on identical station-days: \model{}
(ensemble mean at the 1\,km pixel of the station), the CAMS forecast and Aurora issued at the same time (end of the start
day), and the climatology of the \nrev{1\,km estimates} at the station pixel (grey dashed; Methods); the bias of \model{}, CAMS and Aurora one day ahead is written
in each panel (\textmu g\,m$^{-3}$). \rev{Read at the station pixel on the same day, the \nrev{1\,km estimates} themselves have an RMSE of 2.4 (\textbf{a}), 2.6 (\textbf{b}), 11.2 (\textbf{c}), 8.5 (\textbf{d}), 6.6 (\textbf{e}) and 36.5\,\textmu g\,m$^{-3}$ (\textbf{f}; Methods).} Over all stations, \model{}'s RMSE four days ahead is 14.4\,\textmu g\,m$^{-3}$,
against 16.4 for the CAMS forecast and 17.3 for Aurora, 12\% (95\% interval 7--19\%) and 17\% (13--22\%) lower. \textbf{g}, All 6,324 stations: \model{} read at its 1\,km pixel and averaged to
0.25$^\circ$ and 0.5$^\circ$ before being read like the baselines (equal resolution); averaged to 0.5$^\circ$, its error one
day ahead is 13.5\,\textmu g\,m$^{-3}$. \textbf{h}, Ratio of the ensemble spread
to the RMSE of the ensemble mean by lead time: the 1\,km model on the validation year 2021 against the \nrev{1\,km estimates} (four
members, every eighth start day), after (solid) and before (dashed) the CRPS training, and \model{} trained in Aurora's protocol
on May--November 2022 against the CAMS analysis (eight members); 1 for a calibrated ensemble.}
\label{fig:fair}
\end{figure}
\clearpage

\begin{figure}[!ht]
\centering\includegraphics[width=\textwidth]{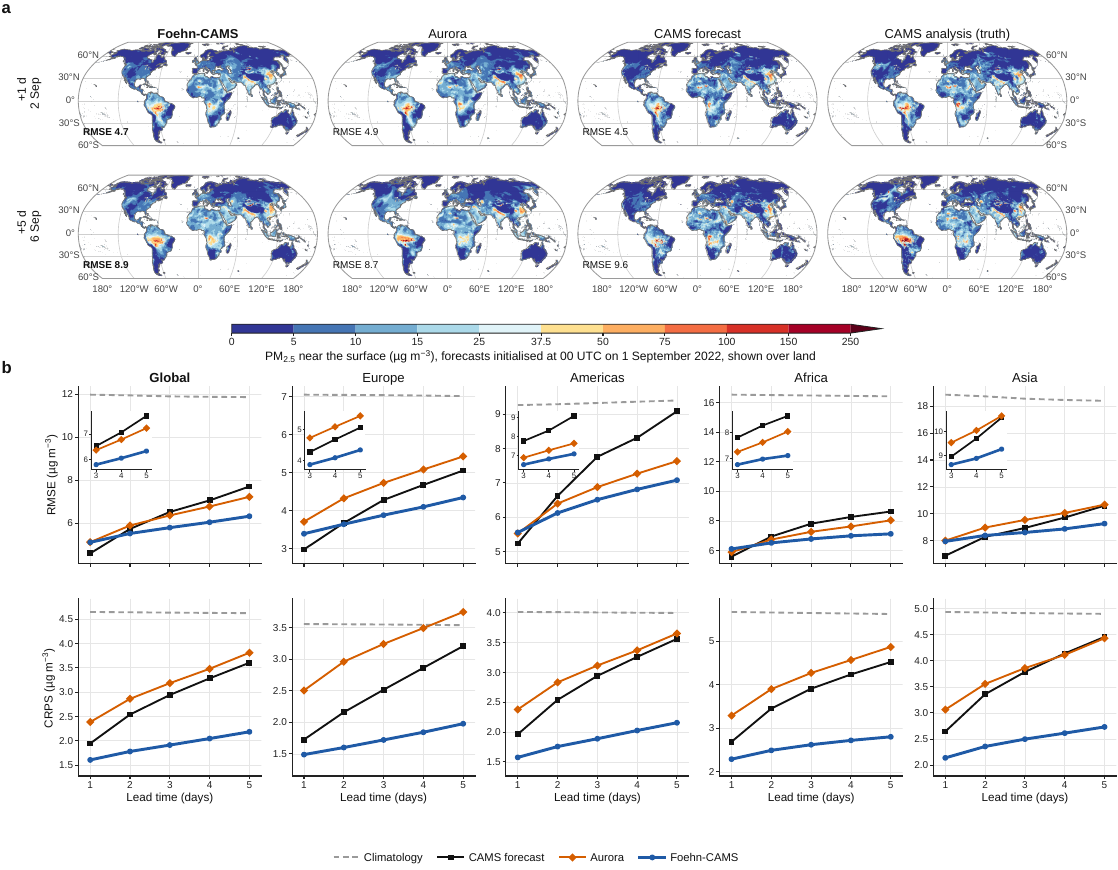}
\caption{\textbf{Trained in Aurora's own setup, Foehn's design has a lower error than Aurora 3 and 5 days ahead, with 131 times fewer parameters.} Foehn-CAMS has Foehn's design (computed transport, learned tracers, flow-matching
network; 9.9 million parameters) but is trained on the 0.4$^\circ$ CAMS analyses with the inputs of Aurora's air-pollution model
(1.3 billion parameters) and forecasts the instantaneous near-surface \pmtf{} every 12\,h up to five days. \textbf{a}, Forecasts
\nrev{initialized} at 00\,UTC on 1 September 2022, one and five days ahead: Foehn-CAMS (ensemble mean of eight members), Aurora run from the
same analyses, the operational CAMS forecast, and the CAMS analysis, the truth of the benchmark; shown over land, scored over the
whole domain (84$^\circ$N--60$^\circ$S; numbers: latitude-weighted RMSE of each map, \textmu g\,m$^{-3}$). \textbf{b}, Latitude-weighted
RMSE (top) and continuous ranked probability score (bottom) against the CAMS analysis by lead time, for the forecasts \nrev{initialized} at
00\,UTC from May to November 2022 (214 \nrev{initializations}) and valid at 00\,UTC, over the whole domain and four continents. The CRPS of
Foehn-CAMS is the fair CRPS of its eight members; for the deterministic Aurora and CAMS forecasts it reduces to the absolute error.
Grey dashed, climatology: the CAMS analyses of 2018--2021 at the same hour within $\pm$15 days of the calendar day (their mean for the
RMSE, the 124 of them as an ensemble for the CRPS). Insets, the RMSE three to five days ahead. Six regions at 12-hour steps, with the
intervals of the differences, are in \edfig~\ref{fig:aurorareg}.}
\label{fig:auroraset}
\end{figure}
\clearpage

\begin{figure}[!ht]
\centering\includegraphics[width=\textwidth]{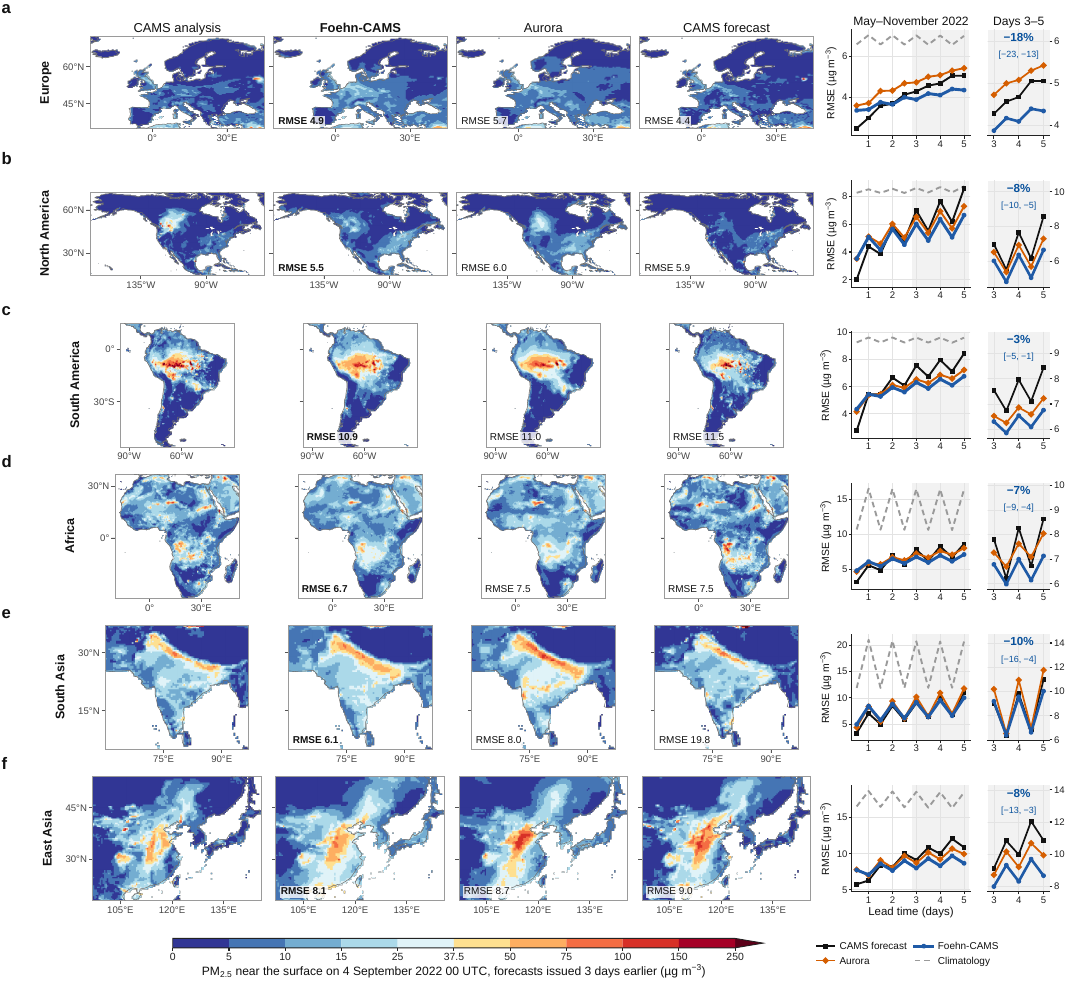}
\caption{\textbf{In Aurora's setup, Foehn's error 3 days ahead is lower than Aurora's over 79\% of the Earth's surface and about half of the land.} \textbf{a}--\textbf{f}, Near-surface \pmtf{} on 4 September 2022 at
00\,UTC forecast three days earlier, over six regions: the CAMS analysis (truth), Foehn-CAMS (ensemble mean), Aurora and the CAMS
forecast, shown over land, with the latitude-weighted RMSE of each map over the region (\textmu g\,m$^{-3}$); right, the
latitude-weighted RMSE of the region by lead time at 12-hour steps over May--November 2022 (214 \nrev{initializations}), with the
climatology of the CAMS analyses of 2018--2021 (grey dashed), and beside it the days three to five (shaded) enlarged, with the change
of Foehn-CAMS's RMSE relative to Aurora three days ahead and its 95\% interval (7-day moving-block bootstrap over \nrev{initializations};
blue when the interval excludes zero). Three days ahead, Foehn-CAMS's error is lower than Aurora's over 79\% of the Earth's
surface but over about half of the land area.}
\label{fig:aurorareg}
\end{figure}

\clearpage
\begin{figure}[!ht]
\centering\includegraphics[width=\textwidth]{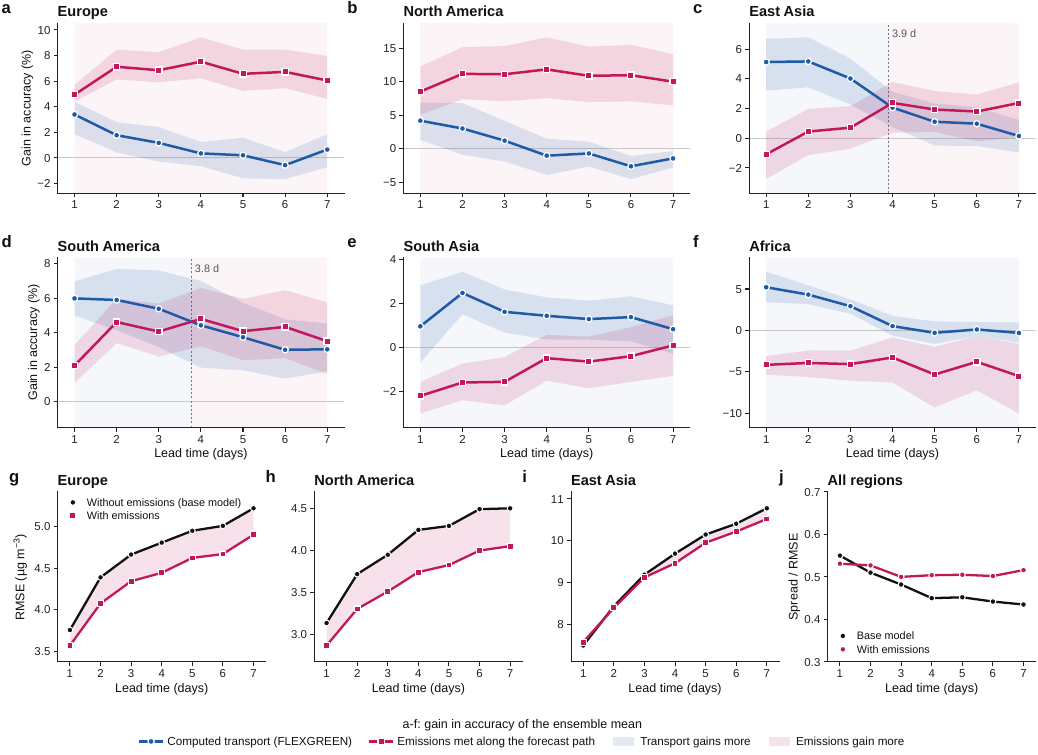}
\caption{\textbf{The computed transport adds skill at short lead times, and \nrev{the emissions taken up along the way} add it at longer ones.} \textbf{a}--\textbf{f}, Gain in the accuracy of the ensemble mean (reduction of the root-mean-square error,
RMSE, against the \nrev{1\,km estimates}, on identical 0.25$^\circ$ land cells and days of 2022) brought by the computed transport (blue: the
base model relative to the same network trained without the transport inputs) and by the emissions \nrev{taken up along the back-trajectories}
(red: the model with the six anthropogenic emission channels relative to the base model trained without them), by lead time, in six of
the nine regions of the evaluation (the other three in Methods); shading, 95\% intervals (paired moving-block bootstrap over start days); the background marks which contribution gains
more, and the dotted line the lead time at which the emissions overtake the transport (East Asia, 3.9 days; South America,
3.8 days). \textbf{g}--\textbf{i}, The RMSE itself, by lead time, for the base model and the model with emissions, in Europe,
North America and East Asia. \textbf{j}, Ratio of the ensemble spread to the RMSE of the ensemble mean over all regions together,
by lead time, for the two models. Both comparisons are between models before the calibration training (Methods).}
\label{fig:emislead}
\end{figure}
\clearpage
\begin{figure}[!ht]
\centering\includegraphics[width=\textwidth]{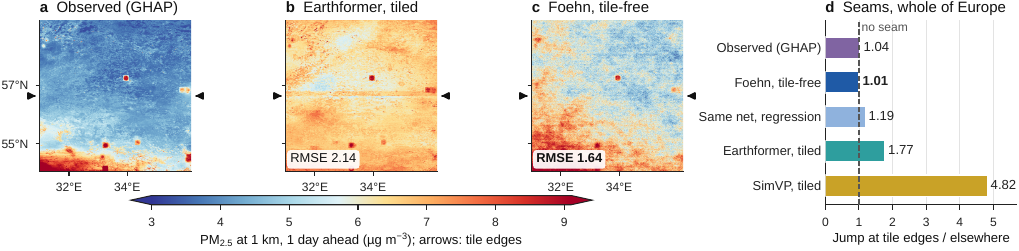}
\caption{\textbf{Over Europe, a network trained at 1\,km and applied tile by tile leaves seams that Foehn's tile-free network avoids.} One-day forecast of \pmtf{} at 1\,km: the \nrev{1\,km estimates}, the target (\textbf{a}), Earthformer trained at 1\,km and applied tile by tile (\textbf{b}) and the tile-free flow-matching network of
Foehn (\textbf{c}); arrows mark tile edges, and the root-mean-square error of each forecast is shown. \textbf{d}, Mean jump of
the forecast across tile edges divided by the jump elsewhere, over Europe (1, no seam).}
\label{fig:km}
\end{figure}
\clearpage


\end{document}